%% file: main.tex
\documentclass[11pt,letterpaper]{mystyle}

\usepackage[comma,authoryear,compress]{natbib}
\usepackage{multirow}
\usepackage{threeparttable}
\usepackage{wrapfig}
\usepackage{pgfplots}
\usepackage[section]{placeins}
\usepackage{dblfloatfix}
\usepackage{mathtools}
\usepackage{mathrsfs}
\usepackage{nicefrac}
\usepackage{dsfont}
\usepackage{microtype}
\usepackage{array}
\usepackage{float}
\tcbuselibrary{most}
\input{math_commands.tex}

\pgfplotsset{compat=1.18}
\definecolor{bestcell}{HTML}{F2F2FE}
\definecolor{mypurple}{HTML}{E6F2EC}

\hypersetup{
    colorlinks=true,
    citecolor={SEUGreenDark},
    linkcolor={SEUGreenDark},
    urlcolor={SEUGreen}
}

\title{\textbf{\textit{Offline Guidance, Online Reasoning:} Reusing LLM Feedback for Small Language Models}}
\runningtitle{Offline Guidance, Online Reasoning: Reusing LLM Feedback for Small Language Models}

\author{Zhang Bohan$^{1}$, Linan Yue$^{1}$, Weibo Gao$^{2}$, Pengyu Chen$^{1}$, Hong Guo$^{1}$, and Yanqi Hao$^{3}$}

\affil[1]{Southeast University}
\affil[2]{Hong Kong Polytechnic University}
\affil[3]{ZTE Corporation}

\correspondingauthor{Linan Yue, Email: \href{lnyue@seu.edu.cn}{lnyue@seu.edu.cn}}

\begin{document}

\begin{abstract}
Large language models (LLMs) offer strong reasoning capabilities but are often costly to access through commercial APIs, while small language models (SLMs) are easier to deploy locally yet remain weaker in reasoning. This capability–deployment gap has motivated LLM–SLM collaboration, which aims to improve SLM reasoning using LLM capabilities while preserving the deployment advantages of SLMs. Existing approaches mainly follow two paradigms. Knowledge distillation uses LLM-generated answers and reasoning trajectories to train SLMs offline, but requires parameter updates and additional training. Alternatively, online collaboration routes difficult problems to an LLM or leverages LLM-generated guidance and corrections when an SLM encounters difficulties. Although effective, online collaboration requires repeated LLM access. Moreover, the guidance produced for a particular problem is discarded after inference and cannot benefit subsequent problems involving similar reasoning states. In the paper, we focus on a more constrained setting in which the LLM is accessed only offline, the SLM parameters remain fixed, and online inference is performed solely by the SLM. To this end, we propose \textbf{Reusable Latent Correction (RLC)}, which converts one-off natural-language guidance from a black-box LLM into persistent corrective experiences in the hidden space of an SLM. RLC stores these experiences in an external bank and retrieves them according to the SLM’s current reasoning state, enabling the SLM to reuse LLM-derived corrections during inference without any online LLM calls. Experiments across multiple reasoning benchmarks and SLM scales show that RLC consistently improves SLM reasoning without parameter updates or online LLM calls. Code is available at \url{https://github.com/ZBH031/reusable-latent-correction}.
\end{abstract}

\maketitle

\section{Introduction}

In recent years, large language models (LLMs) have made remarkable progress on tasks such as mathematical reasoning \citep{hendrycks2021math,yang2024qwen25math} and complex question answering \citep{rein2024gpqa}. However, strong reasoning capabilities typically rely on frontier LLMs at massive scales \citep{openai2023gpt4,guo2025deepseekr1nature,yang2025qwen3}. Most of these LLMs are accessible only through commercial APIs, leaving their parameters and internal states inaccessible. By contrast, small language models (SLMs) with 3B--14B parameters offer flexible deployment, low inference costs, and access to their internal states. Nevertheless, due to their limited model capacity, their reasoning capabilities still fall behind those of frontier LLMs. It is therefore of practical importance to investigate how LLMs can be leveraged to improve the reasoning capabilities of SLMs while preserving the latter's deployment advantages.

\begin{figure*}[t]
    \centering
    \includegraphics[width=14.cm]{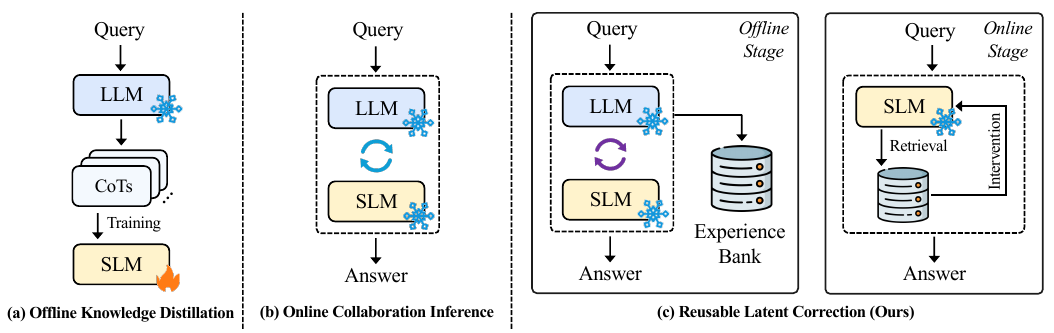}
    \caption{
        Comparison of LLM--SLM collaboration paradigms.
\textbf{(a)} Offline knowledge distillation trains the SLM with LLM-generated CoT trajectories, requiring parameter updates.
\textbf{(b)} Online collaboration avoids retraining but repeatedly queries the LLM during inference.
\textbf{(c)} RLC uses the LLM only for offline experience construction and stores effective corrections in an external Bank. At inference time, the frozen SLM retrieves relevant corrections and applies hidden-state interventions without further LLM access.
    }
    \label{fig:paradigm_comparison}
\end{figure*}

To this end, various paradigms for collaboration between LLMs and SLMs have been proposed. As shown in Figure \ref{fig:paradigm_comparison}, these approaches mainly follow two directions. The first employs knowledge distillation (Figure \ref{fig:paradigm_comparison}(a)), using answers and chain-of-thought (CoT) trajectories generated by an LLM to train an SLM \textbf{offline}, thereby transferring part of the teacher's knowledge to the student \citep{rw_magister2023teaching,rw_hsieh2023distilling,rw_zhu2024pad}. However, such methods typically require updating the parameters of the SLM, which introduces additional training costs. As shown in Figure \ref{fig:paradigm_comparison}(b), the second direction introduces model routing or LLM--SLM collaboration during \textbf{online inference}: the system delegates difficult problems to an LLM or requests guidance and corrections from it when the SLM fails \citep{rw_chen2023frugalgpt,rw_ong2025routellm,lee2025steer,zeng2026glimprouter,rw_kim2025smart}. Although these methods can directly exploit the reasoning capabilities of LLMs, they depend on continued online access to them. Besides, the guidance provided by the LLM is typically used only for the current inference instance. Once the task is completed, this valuable guidance disappears with the context and cannot benefit subsequent problems that encounter similar reasoning states.

These two paradigms exploit LLM capabilities by either updating the SLM’s parameters or repeatedly invoking the LLM. In practical applications, however, we would like to avoid both costs. Specifically, we consider a constrained collaboration setting involving a black-box LLM accessible only through an API and capable of providing natural-language feedback, together with a locally deployable SLM whose internal states are accessible. Our goal is to invoke the LLM only during an offline stage and, without updating the parameters of the SLM, transform its guidance into persistent reasoning experiences. During online inference, the SLM should then solve problems independently, without further access to the LLM. This raises a central research~question:

\textit{Can natural-language guidance provided by a black-box LLM for a specific reasoning error be transformed into an internal corrective experience that an SLM can reuse on future problems?}

To this end, as shown in Figure \ref{fig:paradigm_comparison}(c), we propose \textbf{Reusable Latent Correction (RLC)}, a framework for improving SLM reasoning by reusing experiences acquired from an LLM. RLC transforms the LLM from an online solver into an offline experience provider and converts its one-off natural-language guidance into persistent corrective experiences in the hidden space of the SLM. These experiences are stored explicitly in an external experience bank. During online inference, RLC retrieves and injects relevant corrective experiences on demand according to the current state of the SLM, without invoking the LLM. Specifically, the framework consists of two stages: offline experience construction and online experience reuse.

\textbf{\textit{During offline experience construction}}, the SLM first generates a step-by-step reasoning process for a given problem. Model uncertainty is then used to identify candidate steps that may cause subsequent reasoning to fail. The current problem, the accepted reasoning prefix, and a candidate wrong step are sent to the LLM, which is asked to provide local guidance. Conditioned on this guidance, the SLM regenerates the current step. By comparing the SLM’s internal representations before and after guidance, RLC extracts a latent correction vector that captures the induced change in reasoning. It retains only effective experiences whose corrected reasoning leads to the correct answer. Each retained experience is stored as a key–value pair, with the reasoning state immediately before the wrong generation as the key and the corresponding correction vector as the value, yielding an experience bank that can be queried in subsequent inference.

\textbf{\textit{During online experience reuse}}, inference is performed entirely by the SLM, with no further access to the LLM. Before each reasoning step, the SLM uses its current hidden state to retrieve relevant experiences from the bank. The retrieved correction is then applied to subsequent generation through a hidden-state intervention. Besides, not every problem requires assistance from external experiences. To reduce unnecessary retrieval and intervention, we further introduce a lightweight router: problems that the SLM can solve independently are handled by the original SLM, and the experience bank is activated only when a correction is likely to be~needed.

We evaluate RLC on multiple reasoning benchmarks using SLMs of different scales. Experimental results show that, without updating any model parameters or invoking the LLM during online inference, RLC consistently improves the reasoning performance of SLMs. For example, RLC improves the average performance of a 7B SLM by \textcolor{black}{7.97 percentage points}, bringing it on par with a 32B SLM. It also improves the performance of a \textcolor{black}{14B} SLM by \textcolor{black}{5.93 percentage points}, achieving results comparable to those of \textcolor{black}{72B LLMs} on several tasks.
\vspace{-0.2cm}
\section{Related Work}
\textbf{Collaborative Reasoning with Large and Small Language Models.} Prior work on combining large and small language models can be broadly divided into distillation-based methods and inference-time collaboration. Reasoning distillation transfers the behavior of a stronger teacher into a smaller student by supervising chain-of-thought rationales \citep{rw_magister2023teaching}, LLM-generated rationales as additional supervision \citep{rw_hsieh2023distilling}, executable programs \citep{rw_zhu2024pad}. Later studies further improve distillation by filtering trajectories \citep{wang-etal-2023-scott,lei-etal-2025-learning}, decomposing tasks \citep{shridhar-etal-2023-distilling}, designing curricula \citep{jiang-etal-2025-teach}, or incorporating symbolic and structured supervision \citep{sun-etal-2024-enhancing-code,liao-etal-2025-skintern}, with broad CoT collections and mentor-based supervision providing additional rationale sources \citep{kim-etal-2023-cot,lee-etal-2024-mentor}. These methods can improve the standalone capability of small models, but adapting to new teacher feedback or new domains typically requires another round of parameter updates.

Inference-time collaboration instead allocates computation between models without fully absorbing the teacher into the student. Query-level systems use cascades \citep{rw_chen2023frugalgpt}, learned routers \citep{rw_ong2025routellm}, taxonomy-based selection \citep{shah-shridhar-2025-select}, or calibrated confidence \citep{zhang-etal-2026-confidence} to decide whether a problem should be handled by a small or large model. More fine-grained methods move the decision inside the reasoning trajectory: AdaSwitch assigns reasoning tokens to different agents \citep{sun-etal-2024-adaswitch}, STEER invokes a large model based on step-level confidence \citep{lee2025steer}, GlimpRouter estimates step difficulty from first-token entropy \citep{zeng2026glimprouter}, TrigReason detects risky reasoning states \citep{rw_zhao2026trigreason}, and token-level routing consults the cloud model only for selected tokens \citep{she-etal-2025-token}. Another line exchanges guidance rather than only routing computation. SMART asks the large model for targeted hints \citep{rw_kim2025smart}, MentorCollab adopts corrective segments under trajectory divergence \citep{rw_wang2026mentorcollab}, Tandem uses the large model as a strategic coordinator \citep{fu-etal-2026-tandem}, LM-Guided CoT studies small-model guidance for black-box large-model reasoning \citep{lee-etal-2024-small}, and PyroDash learns control tokens for generation-time handoff \citep{lyu2026pyrodash}. These approaches reduce unnecessary large-model computation, but their guidance is usually consumed only by the current instance. In contrast, RLC converts effective offline teacher guidance into persistent latent corrections stored in an external Bank, enabling a frozen small model to reuse prior guidance during future reasoning without online large-model calls or parameter updates.

\textbf{Inference-Time Activation Steering for Reasoning.} Activation steering controls generation by modifying hidden representations during inference. Early methods construct steering directions from contrastive prompt pairs \citep{rw_turner2023actadd}, multiple positive--negative examples \citep{rw_rimsky2024caa}, or in-context demonstrations \citep{rw_liu2024icv}, providing a lightweight alternative to fine-tuning. Subsequent work studies more controlled intervention operations, including scaling selected activation dimensions \citep{stoehr-etal-2024-activation}, pseudo-rotating hidden states while preserving norms \citep{pham-nguyen-2024-householder}, and constraining intervened states to remain close to the model's original activation distribution \citep{vogels-etal-2026-distribution}. Activation steering has also been applied to specific behaviors such as debiasing \citep{li-etal-2025-fairsteer}, chain-of-thought compression \citep{azizi-etal-2026-activation}, and cross-lingual reasoning transfer \citep{maraia-etal-2026-activation}.

Recent reasoning-oriented steering methods make interventions more adaptive. ReBalance adjusts steering based on real-time confidence \citep{li2026rebalance}, PDS retrieves input-matched reasoning prototypes \citep{kayan2025pds}, K-CAST estimates intervention coefficients from neighboring examples \citep{rw_valentino2026kcast}, and RISER composes latent reasoning-skill vectors with a lightweight router \citep{rw_ye2026riser}. However, these methods mainly rely on global directions, input-level prototypes, or reusable skills, limiting their ability to match the local states that emerge during multi-step reasoning. In contrast, RLC stores state-conditioned latent corrections from local teacher guidance and retrieves them according to the evolving reasoning state, enabling step-specific intervention and abstention when no relevant experience is found.


\begin{figure}[t]
    \centering
    \includegraphics[width=14.cm]{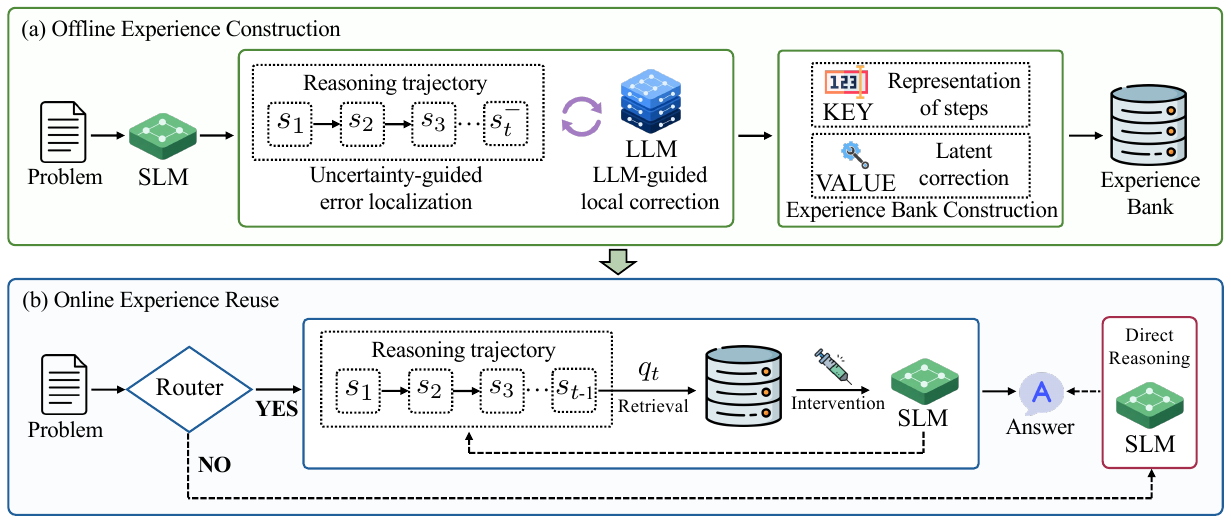}
    \caption{
    Overall architecture of Reusable Latent Correction (RLC), which consists of offline experience construction and online experience reuse.
    }
    \label{fig:rlc_overview}
    \vspace{-0.4cm}
\end{figure}
\section{Problem Formulation}
\label{sec:problem}
In this paper, we consider a constrained LLM--SLM collaboration setting consisting of a locally deployable small language model (SLM) $M_S$ and a black-box large language model (LLM) $M_L$. Among them, the parameters and hidden states of $M_S$ are accessible, whereas $M_L$ can only be queried through an API.  The LLM is available during offline preparation but cannot be accessed during online inference. Moreover, the parameters of the SLM is fixed throughout the entire process.

Let $\mathcal{D}_{\mathrm{off}}=\{(x_i,a_i)\}_{i=1}^{N}$ denote a set of offline reasoning problems with reference answers. Our objective consists of two stages. During the offline stage, we use natural-language guidance obtained from $M_L$ on $\mathcal{D}_{\mathrm{off}}$ to construct an external experience bank $\mathcal{B}$ without updating the parameters of $M_S$. During online inference, given a problem $x \in \mathcal{D}_{\mathrm{on}}$, the SLM retrieves relevant corrections from $\mathcal{B}$ according to its evolving reasoning states and uses them to guide subsequent generation. Therefore, online inference is performed solely by $M_S$ with the offline-constructed bank $\mathcal{B}$, without further access to $M_L$. The overall process can be formulated as:
\begin{equation}
\mathcal{B}
=
\mathcal{A}_{\mathrm{off}}
\left(
M_S,M_L,\mathcal{D}_{\mathrm{off}}
\right), \quad
\widehat{a}
=
\mathcal{A}_{\mathrm{on}}
\left(
M_S,x,\mathcal{B}
\right),
\label{eq:problem}
\end{equation}
where $\mathcal{A}_{\mathrm{off}}$ denotes the offline procedure for constructing the experience bank, $\mathcal{A}_{\mathrm{on}}$ is the online inference procedure for reusing the stored experiences.


\vspace{-0.3cm}
\section{Reusable Latent Correction}
\vspace{-0.3cm}
\label{sec:method}
As shown in Figure~\ref{fig:rlc_overview}, we propose \textbf{Reusable Latent Correction (RLC)}, which converts one-off LLM guidance into reusable latent corrections for a fixed SLM. During \textbf{offline construction}, uncertain reasoning steps are corrected with LLM guidance and stored as KEY--VALUE experiences. During \textbf{online reuse}, the SLM retrieves and injects relevant corrections based on its current reasoning state, enabling improved reasoning without online LLM access.

\vspace{-0.2cm}
\subsection{Offline Experience Construction}
\vspace{-0.2cm}
\label{sec:offline}

Given a problem $x \in \mathcal{D}_{\mathrm{off}}$, the SLM generates a reasoning trajectory $s=(s_1,\ldots,s_T)$ step by step, where each step is separated by a predefined delimiter. Before generating step $t$, the current context is $c_t=(x,s_{1:t-1})$, and the next step is generated as:
\begin{equation}
s_t
\sim
M_S
\left(
\cdot \mid c_t
\right),
\qquad
c_t=(x,s_{1:t-1}).
\label{eq:step_generation}
\end{equation}
The offline stage then identifies uncertain steps, obtains local LLM guidance, and converts successful corrections into reusable latent experiences.


\textbf{Uncertainty-guided error localization.} Given the reasoning trajectory $s$, we aim to identify steps at which the SLM has not formed a stable prediction. Since $M_S$ consists of multiple Transformer layers, the representation and prediction of each token are progressively refined from intermediate layers to the final layer. For a confident reasoning step, predictions from intermediate layers are expected to be consistent with the final prediction, whereas large disagreement indicates greater uncertainty \citep{chuang2024dola,lin2026controlling}. We therefore estimate step-level uncertainty by measuring the divergence between selected intermediate-layer predictions and the final-layer prediction. Let $P_t$ denote the token positions belonging to step $s_t$, and let $h_\tau^l$ denote the hidden state at token position $\tau$ in layer $l$. For each selected intermediate layer $l\in\mathcal{L}$, we project its normalized hidden state into the vocabulary space:
\begin{equation}
p_l^{(\tau)}
=
\operatorname{softmax}
\left(
\frac{
W_E\operatorname{LN}(h_\tau^l)
}{
T_1
}
\right),
\qquad
u_t
=
\frac{1}{|\mathcal{L}||P_t|}
\sum_{l\in\mathcal{L}}
\sum_{\tau\in P_t}
D_{\mathrm{JS}}
\left(
p_N^{(\tau)}
\parallel
p_l^{(\tau)}
\right).
\label{eq:uncertainty}
\end{equation}
where $W_E$ is the output projection matrix, $p_N^{(\tau)}$ is the corresponding final-layer distribution, and $T_1$ is a temperature parameter. A larger $u_t$ indicates stronger disagreement between intermediate- and final-layer predictions. We identify $s_t^{-}$ as an uncertain reasoning step when $u_t>\tau_{\mathrm{unc}}$, where $\tau_{\mathrm{unc}}$ is a predefined threshold that determines whether LLM guidance is triggered for the current step.



\textbf{LLM-guided local correction.} When an uncertain step $s_t^{-}$ is detected, RLC provides the LLM with the problem $x$ and the partial reasoning trajectory $s_{1:t}$, and asks it to diagnose potential issues in the uncertain step and provide targeted natural-language guidance $g_t$ for correcting that step:
\begin{equation}
g_t
=
M_L
\left(
x,s_{1:t-1},s_t^{-}
\right).
\label{eq:local_correction1}
\end{equation}
Then, the returned guidance $g_t$ is appended to the original reasoning context $c_t$ to form the guided input $\widetilde{c}_t$, which is fed into the SLM to regenerate the current step:
\begin{equation}
\widetilde{c}_t
=
c_t \oplus g_t
=
\left(
x,s_{1:t-1},g_t
\right),
\qquad
s_t^{+}
\sim
M_S
\left(
\cdot\mid\widetilde{c}_t
\right),
\label{eq:local_correction}
\end{equation}
where $\oplus$ denotes textual concatenation. Once $s_t^{+}$ is generated, the guidance $g_t$ is removed, and subsequent reasoning proceeds from the corrected context $c_{t+1}=(x,s_{1:t-1},s_t^{+})$. Then, the same uncertainty estimation and local correction are repeatedly applied to subsequent reasoning steps until the complete reasoning trajectory is generated.

\textbf{Experience Bank Construction.} After obtaining LLM guidance, we aim to preserve its effect on the SLM rather than the guidance text itself. To this end, RLC stores each candidate experience as a KEY--VALUE pair $(k_t,v_t)$. The key $k_t$ records the SLM reasoning state immediately before the uncertain step is generated and serves as the retrieval representation for matching similar future states, while the value $v_t$ captures the change in the SLM hidden space induced by the LLM guidance.

Before describing the construction of the keys and values, we first introduce the hidden-state representations used by RLC. Let $\ell^*$ denote the layer used for experience retrieval and hidden-state intervention. For any input sequence $z$, $h_{\mathrm{last}}^{\ell^*}(z)$ denotes the hidden state of its final token at layer~$\ell^*$. For a reasoning step $s_j$, we define its mean representation as $ \bar{h}^{\ell^*}(s_j)=\frac{1}{|s_j|}\sum_{\tau\in s_j}h_\tau^{\ell^*}$, where $h_\tau^{\ell^*}$ denotes the hidden state of token $\tau$ at layer $\ell^*$.

\textbf{\textit{(i) Construction of the retrieval KEY.}} Based on these representations, we construct the retrieval key $k_t$ as follows. For the first reasoning step, the key is the hidden state of the final token in the problem prompt. For a later step, the key is the mean representation of the preceding reasoning step $s_{t-1}$:
\begin{equation}
k_t=\mathbb{I}[t=1]\,h_{\mathrm{last}}^{\ell^*}(x)+\mathbb{I}[t>1]\,\bar{h}^{\ell^*}(s_{t-1}),
\label{eq:key}
\end{equation}
where $\mathbb{I}[\cdot]$ is the indicator function. The key therefore describes the SLM state immediately before step $s_t$ is generated, without containing information from the uncertain step itself. Because the same state is available during online inference, RLC can retrieve a relevant correction before a similar uncertain step is generated.

\textbf{\textit{(ii) Construction of the VALUE.}} We use the latent change induced by LLM guidance as the value~$v_t$ of each experience. Specifically, the correction vector contains two complementary components. The \emph{process correction} $v_t^{\mathrm{proc}}$ captures how adding the guidance changes the generation-boundary state before the SLM regenerates the current step. The \emph{result correction} $v_t^{\mathrm{result}}$ captures the representational difference between the original step $s_t^{-}$ and the guided step $s_t^{+}$. They are defined as:
\begin{equation}
v_t^{\mathrm{proc}}
=
h_{\mathrm{last}}^{\ell^*}
\left(
\widetilde{c}_t
\right)
-
h_{\mathrm{last}}^{\ell^*}
\left(
c_t
\right), \quad
v_t^{\mathrm{result}}
=
\bar{h}^{\ell^*}
\left(
s_t^{+}
\right)
-
\bar{h}^{\ell^*}
\left(
s_t^{-}
\right), \quad
v_t
=
\left[
v_t^{\mathrm{proc}};
v_t^{\mathrm{result}}
\right].
\label{eq:correction_vector}
\end{equation}
Finally, all KEY--VALUE pairs are collected into the experience bank $\mathcal{B}=\{(k_{t_j},v_{t_j})\}_{j=1}^{m}$, where $m$ is the number of retained experiences. It is noting that a candidate correction is retained only if continuing the reasoning process from the regenerated step eventually yields the correct answer.~This filtering prevents locally plausible but globally ineffective corrections from being stored in the bank.




\subsection{Online Experience Reuse}
\label{sec:online}

Given a problem $x$, online inference is performed using only the SLM and the offline-constructed experience bank $\mathcal{B}$, without further access to the LLM. The SLM generates a reasoning trajectory $s=(s_1,\ldots,s_T)$ step by step. Before generating each step, RLC constructs a query from the current reasoning state, retrieves and aggregates relevant corrections from $\mathcal{B}$, and then applies the aggregated correction to the subsequent generation. This process consists of two components: state-conditioned retrieval and aggregation, followed by two-part hidden-state intervention.

\textbf{State-conditioned Retrieval and Aggregation.} After completing step $s_{t-1}$ and before generating step $s_t$, RLC constructs a query $q_t$ using the same representation employed for the offline keys (i.e., $q_t=\mathbb{I}[t=1]\,h_{\mathrm{last}}^{\ell^*}(x)+\mathbb{I}[t>1]\,\bar{h}^{\ell^*}(s_{t-1})$). Using the same representation for offline keys and online queries ensures that the current reasoning state can be matched against the stored experiences. For each bank item $(k_i,v_i)\in\mathcal{B}$, we compute its cosine similarity to the current query as:
\begin{equation}
\rho_{t,i}
=
\cos(q_t,k_i)
=
\frac{
q_t^\top k_i
}{
\lVert q_t\rVert_2
\lVert k_i\rVert_2
}.
\end{equation}
Here, $\rho_{t,i}$ measures the relevance of the $i$-th stored experience to the reasoning state before step $s_t$. If the highest similarity is below the threshold $\tau_{\mathrm{sim}}$, RLC treats the retrieval as unsuccessful, sets $\mathcal{N}_t=\varnothing$, and generates the current step without correction. Otherwise, the indices of the $K$ most similar experiences form the retrieved set, i.e., $\mathcal{N}_t=\operatorname{TopK}_{i\in\{1,\ldots,|\mathcal{B}|\}}(\rho_{t,i})$ when $\max_i\rho_{t,i}\geq\tau_{\mathrm{sim}}$. When $\mathcal{N}_t\neq\varnothing$, RLC aggregates the correction vectors of the retrieved experiences into a single correction applicable to the current reasoning state. Specifically, the normalized similarity weight of experience $i$ is $\omega_{t,i}=\exp(\rho_{t,i}/T_{\mathrm{ret}})\big/\sum_{j\in\mathcal{N}_t}\exp(\rho_{t,j}/T_{\mathrm{ret}})$, \textcolor{black}{and the aggregated correction is $\bar{v}_t=\sum_{i\in\mathcal{N}_t}\omega_{t,i}v_i=[\bar{v}_t^{\mathrm{proc}};\bar{v}_t^{\mathrm{result}}]$}. Here, $T_{\mathrm{ret}}$ controls the concentration of the weights. The resulting process and result components are then applied at different stages of SLM generation.


\textbf{Two-part Hidden-state Intervention.} The aggregated correction contains two components that intervene at different stages of generating reasoning step $s_t$. First, after prefilling the current context $(x,s_{1:t-1})$, we inject the process correction into the last-token hidden state at layer $\ell^*$:
\begin{equation}
\widetilde{h}_{t,\mathrm{last}}^{\ell^*}
=
h_{t,\mathrm{last}}^{\ell^*}
+
\alpha_p\bar{v}_t^{\mathrm{proc}}.
\label{eq:process_injection}
\end{equation}
This intervention steers the overall reasoning direction before the next step is generated. Then, during autoregressive decoding, we inject the result correction into the hidden state of every token position $\tau$ belonging to the generated step:
\begin{equation}
\widetilde{h}_{\tau}^{\ell^*}
=
h_{\tau}^{\ell^*}
+
\alpha_r\bar{v}_t^{\mathrm{result}},
\qquad \tau\in P_t.
\label{eq:result_injection}
\end{equation}
where $P_t$ denotes the set of token positions in $s_t$, and $\alpha_p$ and $\alpha_r$ control the strengths of the process and result interventions, respectively.

The process correction shifts the initial direction of the upcoming reasoning step, while the result correction continuously guides its token-level generation. RLC repeats retrieval and intervention before each subsequent step until the SLM produces the final answer. Because retrieval is conditioned on the state preceding a potential error, historical LLM guidance can be reused proactively without online LLM access.

\textbf{Problem-level Router.} In practice, bank augmentation is not always necessary: the original SLM can solve some problems independently, while irrelevant retrieved experiences may introduce unnecessary corrections. We therefore employ a lightweight problem-level router $R_\phi$ to determine whether the experience bank should be activated based solely on the input problem $x$, i.e., $r=\mathbb{I}\!\left[R_\phi(x)\geq\tau_R\right]$, where $\tau_R$ is the routing threshold. When $r=0$, the original SLM directly solves the problem. When $r=1$, step-level experience retrieval and hidden-state intervention are activated. The construction and training details of the router are provided in Appendix~\ref{app:router}.
Besides, we evaluate whether the Router can effectively determine when corrective intervention is needed, with detailed results reported in Appendix~\ref{sec:rq2}.

The complete procedure of RLC is summarized in Algorithm~\ref{alg:rlc} in Appendix~\ref{Algorithm}.

\section{Experiments}
\vspace{-0.1cm}
\label{sec:experiments}
To comprehensively evaluate RLC, we investigate the following seven research questions: \textbf{RQ1:} How effectively does RLC improve the reasoning performance of SLMs across different model scales? \textbf{RQ2:} Can latent corrections constructed from one model scale generalize to models of different scales? \textbf{RQ3:} Can reusable latent corrections provide comparable effectiveness to directly obtaining LLM guidance during inference? \textbf{RQ4:} Does RLC preserve the inference efficiency of the underlying SLM while improving its reasoning performance?

\vspace{-0.2cm}
\subsection{Benchmarks and Experimental Setups}
\label{sec:benchmarks_models}
\textbf{Benchmarks.} We evaluate RLC on five reasoning benchmarks: MATH500 \citep{lightman2023letsverifystepstep}, GSM8K \citep{cobbe2021trainingverifierssolvemath}, AMC24, AIME \citep{hendrycks2021math}, and GPQA-Diamond \citep{rein2024gpqa}. For offline Bank construction, we use the full GSM8K and MATH training sets, totaling 14,973 problems. These examples are used to generate reasoning trajectories, identify errors, and extract latent corrections from localized LLM guidance. No test examples are used during Bank construction, ensuring separation from downstream evaluation. Details are shown in Appendix~\ref{app:benchmarks}.

\textbf{Baselines.} We compare RLC with two categories of baselines. The first is \emph{LLM--SLM collaborative inference}, including STEER \citep{lee2025steer} and GlimpRouter \citep{zeng2026glimprouter}, which invoke an LLM during test-time reasoning. The second is \emph{inference-time representation intervention}, including ReBalance \citep{li2026rebalance} and PDS \citep{kayan2025pds}, which steer SLM reasoning by modifying hidden representations using signals derived from the SLM itself. In contrast, RLC uses LLM guidance only offline and reuses the resulting latent corrections during online SLM inference. Detailed baseline descriptions are provided in Appendix~\ref{app:experimental_details}.

\textbf{Experimental Setups.} We use Qwen2.5-3B/7B/14B-Instruct as locally deployed SLMs and Qwen3-Max as the black-box LLM only for offline experience construction. All SLM parameters remain frozen. Qwen2.5-7B/32B/72B Native are included only as standalone reference models and are not used as teachers. Unless otherwise specified, RLC retrieves the top-3 experiences with $\tau_{\mathrm{sim}}=0.95$, uses $(\alpha_p,\alpha_r)=(0.5,0.25)$, and applies retrieval and intervention at the fourth Transformer layer from the output. We report \textbf{accuracy (Acc.)} as the primary metric, with \textbf{Tok.} denoting the average generated tokens per problem. Additional details are provided in Appendix~\ref{app:experimental_details}.
\suppressfloats[t]
\begin{table*}[!t]
\centering
\begingroup
\caption{
Main results across five reasoning benchmarks. Among them, for STEER and GlimpRouter, $A\rightarrow B$ denotes that model $A$ is assisted by model $B$ during inference. Best and second-best results are highlighted in bold and underline, respectively.}
\vspace{0.1cm}
\label{tab:main_results}
\endgroup
\renewcommand{\arraystretch}{.75}
\setlength{\tabcolsep}{2.2mm}
      \scalebox{0.77}{
\begin{tabular}{@{}l*{6}{rr}@{}}
\toprule
\multirow{2}{*}{Method} & \multicolumn{2}{c}{{MATH500}} & \multicolumn{2}{c}{{GSM8K}} & \multicolumn{2}{c}{{GPQA}} & \multicolumn{2}{c}{{AMC24}} & \multicolumn{2}{c}{{AIME25}} & \multicolumn{2}{c}{{Average}} \\
\cmidrule(lr){2-3}\cmidrule(lr){4-5}\cmidrule(lr){6-7}\cmidrule(lr){8-9}\cmidrule(lr){10-11}\cmidrule(lr){12-13}
& Acc. & Tok. & Acc. & Tok. & Acc. & Tok. & Acc. & Tok. & Acc. & Tok. & Acc. & Tok. \\
\midrule
Qwen2.5-3B Native & 67.00 & 607.84 & 84.69 & 299.69 & 30.30 & 740.65 & 4.44 & 967.53 & 3.33 & 888.30 & 37.95 & 700.80 \\
Qwen2.5-7B Native & \textbf{74.20} & \textbf{557.86} & \textbf{90.83} & 290.12 & 38.89 & \textbf{586.81} & 20.00 & 1030.87 & \textbf{10.00} & 931.47 & \underline{46.78} & 679.43 \\
ReBalance (3B) & 70.60 & 619.74 & 86.73 & 290.96 & \underline{39.45} & 724.48 & 22.22 & \textbf{764.89} & 3.33 & \textbf{870.37} & 44.47 & \textbf{654.09} \\
PDS (3B) & 71.80 & 636.74 & 86.58 & 295.21 & 33.84 & 700.39 & 22.22 & 963.78 & \underline{6.67} & 1304.10 & 44.22 & 780.04 \\
STEER (3B$\rightarrow$7B) & 68.80 & 626.16 & 87.64 & 346.59 & 32.32 & 737.54 & \textbf{24.44} & 857.69 & \textbf{10.00} & \underline{872.47} & 44.64 & 688.09 \\
GlimpRouter (3B$\rightarrow$7B) & 68.40 & 594.96 & 84.38 & \underline{281.51} & 32.32 & \underline{638.50} & \textbf{24.44} & \underline{853.87} & 0.00 & 928.47 & 41.91 & \underline{659.46} \\
\midrule
\rowcolor{mypurple}
\textbf{RLC (3B)} & \underline{73.00} & \underline{576.91} & \underline{89.99} & \textbf{272.68} & \textbf{40.91} & 690.92 & \textbf{24.44} & 992.78 & \textbf{10.00} & 921.93 & \textbf{47.67} & 691.04 \\
Prefill only & 71.80 & 571.79 & 89.99 & 272.29 & 40.91 & 690.01 & 24.44 & 1420.96 & 3.33 & 1106.17 & 46.09 & 812.24 \\
Decode only & 69.80 & 605.78 & 89.69 & 273.01 & 40.92 & 691.54 & 20.00 & 1326.51 & 3.33 & 1148.33 & 44.75 & 809.03 \\
\bottomrule
\end{tabular}}

\vspace{1.0mm}
\setlength{\tabcolsep}{2.2mm}
      \scalebox{0.76}{
\begin{tabular}{@{}l*{6}{rr}@{}}
\toprule
\multirow{2}{*}{Method} & \multicolumn{2}{c}{{MATH500}} & \multicolumn{2}{c}{{GSM8K}} & \multicolumn{2}{c}{{GPQA}} & \multicolumn{2}{c}{{AMC24}} & \multicolumn{2}{c}{{AIME24}} & \multicolumn{2}{c}{{Average}} \\
\cmidrule(lr){2-3}\cmidrule(lr){4-5}\cmidrule(lr){6-7}\cmidrule(lr){8-9}\cmidrule(lr){10-11}\cmidrule(lr){12-13}
& Acc. & Tok. & Acc. & Tok. & Acc. & Tok. & Acc. & Tok. & Acc. & Tok. & Acc. & Tok. \\
\midrule
Qwen2.5-7B Native & 74.20 & 557.86 & 90.83 & 290.12 & 38.89 & 586.81 & 20.00 & 1030.87 & 13.33 & 1398.97 & 47.45 & 772.93 \\
Qwen2.5-32B Native & \textbf{83.20} & \textbf{530.40} & \textbf{95.30} & \textbf{231.24} & \underline{42.93} & \textbf{481.80} & \textbf{37.78} & 947.33 & 13.33 & \textbf{839.73} & \underline{54.51} & \textbf{606.10} \\
ReBalance (7B) & 79.20 & 592.87 & 91.89 & 316.16 & 40.40 & 728.55 & \underline{28.89} & 894.62 & 13.33 & 1002.87 & 50.74 & 707.01 \\
PDS (7B) & 76.80 & 597.08 & 91.28 & 305.93 & 40.91 & 703.36 & 26.67 & \textbf{791.38} & \underline{16.67} & 1259.30 & 50.47 & 731.41 \\
STEER (7B$\rightarrow$32B) & 77.00 & 596.10 & 91.21 & 309.08 & 40.91 & 748.68 & 22.22 & \underline{887.07} & 10.00 & \underline{863.27} & 48.27 & \underline{680.84} \\
GlimpRouter (7B$\rightarrow$32B) & 77.40 & 589.24 & 90.90 & \underline{270.03} & 41.92 & 665.14 & 26.67 & 987.84 & \textbf{20.00} & 956.40 & 51.38 & 693.73 \\
\midrule
\rowcolor{mypurple}
\textbf{RLC (7B)} & \underline{79.80} & 564.70 & \underline{94.09} & 279.07 & \textbf{45.45} & \underline{654.44} & \textbf{37.78} & 1100.38 & \textbf{20.00} & 1300.53 & \textbf{55.42} & 779.82 \\
Prefill only & 76.80 & 577.16 & 94.01 & 278.70 & 44.95 & 653.01 & 26.67 & 914.67 & 13.33 & 1329.37 & 51.15 & 750.58 \\
Decode only & 75.40 & 577.25 & 94.09 & 278.82 & 45.45 & 649.17 & 28.89 & 1143.00 & 10.00 & 1187.13 & 50.77 & 767.07 \\
\bottomrule
\end{tabular}}

\vspace{1.0mm}
\setlength{\tabcolsep}{2.2mm}
      \scalebox{0.75}{
\begin{tabular}{@{}l*{6}{rr}@{}}
\toprule
\multirow{2}{*}{Method} & \multicolumn{2}{c}{{MATH500}} & \multicolumn{2}{c}{{GSM8K}} & \multicolumn{2}{c}{{GPQA}} & \multicolumn{2}{c}{{AMC24}} & \multicolumn{2}{c}{{AIME24}} & \multicolumn{2}{c}{{Average}} \\
\cmidrule(lr){2-3}\cmidrule(lr){4-5}\cmidrule(lr){6-7}\cmidrule(lr){8-9}\cmidrule(lr){10-11}\cmidrule(lr){12-13}
& Acc. & Tok. & Acc. & Tok. & Acc. & Tok. & Acc. & Tok. & Acc. & Tok. & Acc. & Tok. \\
\midrule
Qwen2.5-14B Native & 77.00 & 602.48 & 94.47 & 290.74 & 42.93 & \textbf{421.87} & 31.11 & \underline{766.82} & 13.33 & 1122.87 & 51.77 & \textbf{640.96} \\
Qwen2.5-72B Native & \textbf{83.80} & 567.37 & \textbf{95.75} & 284.14 & \textbf{52.02} & 618.53 & \textbf{40.00} & 865.44 & \underline{16.67} & 1169.13 & \underline{57.65} & 700.92 \\
ReBalance (14B) & 79.40 & 624.00 & 94.92 & 320.64 & 46.46 & 651.65 & 35.56 & 1230.49 & 10.00 & 1440.00 & 53.27 & 853.36 \\
PDS (14B) & 81.00 & 602.52 & 94.92 & 310.04 & \underline{47.98} & 703.52 & \textbf{40.00} & \textbf{762.51} & \underline{16.67} & \underline{1031.60} & 56.11 & 682.04 \\
STEER (14B$\rightarrow$72B) & 79.80 & 627.11 & 95.22 & 342.72 & 45.96 & 798.60 & 31.11 & 886.20 & 10.00 & \textbf{1025.50} & 52.42 & 736.03 \\
GlimpRouter (14B$\rightarrow$72B) & 77.60 & \textbf{567.27} & 94.84 & \textbf{260.38} & 45.45 & 653.32 & 26.67 & 830.44 & 13.33 & 1158.23 & 51.58 & \underline{693.93} \\
\midrule
\rowcolor{mypurple}
\textbf{RLC (14B)} & \underline{83.00} & \underline{583.87} & \underline{95.68} & \underline{280.45} & \textbf{52.02} & \underline{467.12} & \underline{37.78} & 864.58 & \textbf{20.00} & 1120.07 & \textbf{57.70} & 663.22 \\
Prefill only & 79.00 & 577.29 & 95.38 & 280.43 & 52.02 & 470.86 & 33.33 & 1128.38 & 20.00 & 1172.20 & 55.95 & 725.83 \\
Decode only & 78.40 & 606.58 & 95.60 & 281.16 & 51.01 & 467.83 & 33.33 & 972.69 & 13.33 & 1163.13 & 54.33 & 698.28 \\
\bottomrule
\end{tabular}}
\end{table*}

\vspace{-0.4cm}
\subsection{Overall Performance (RQ1)}
\label{sec:rq1}
\begin{wrapfigure}{r}{0.45\columnwidth}
    \centering
    \vspace{-0.2cm}
    \includegraphics[width=5cm]{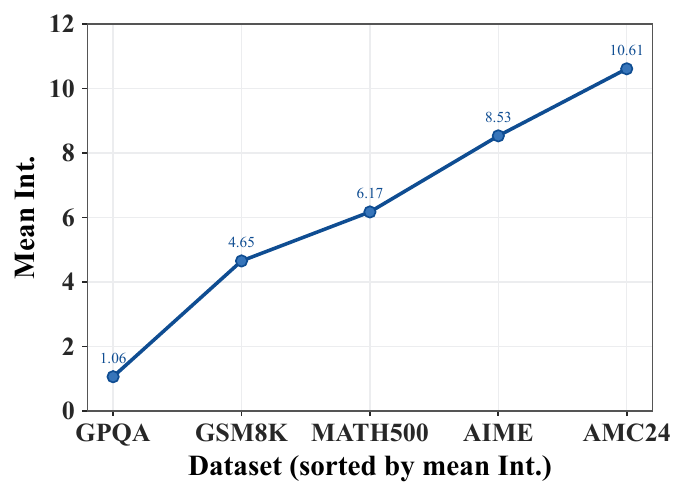}
    \vspace{-0.4cm}
\caption{
Intervention behavior across tasks.
Int. denotes the average number of interventions among Bank-routed problems, averaged across the RLC (3B, 7B, and 14B).
}
    \label{fig:int-profile}
\end{wrapfigure}

To evaluate the overall effectiveness of RLC, we conduct experiments on five reasoning benchmarks, with results shown in Table~\ref{tab:main_results} and Figure~\ref{fig:int-profile}. RLC consistently improves SLM reasoning across model scales and substantially narrows the gap to larger models. In particular, RLC (3B) and RLC (7B) slightly outperform the native 7B and 32B models, respectively, while RLC (14B) achieves performance comparable to the native 72B model. Meanwhile, its output length remains close to that of the corresponding Native models, suggesting that the gains mainly come from more effective reasoning correction rather than longer trajectories. Besides, we have the following observations:

\textbf{Comparison with baselines.} Compared with \textit{representation-intervention baselines}, RLC consistently achieves stronger performance across model scales. The key difference is that existing methods mainly derive generic steering signals from the model's own representations, whereas RLC constructs error-specific corrections from actual reasoning failures and offline LLM guidance. Each correction is further tied to its originating reasoning state, enabling state-adaptive intervention as the trajectory evolves. RLC also compares favorably with \textit{LLM--SLM collaborative inference methods}. While these approaches require online LLM participation, RLC reuses offline guidance and performs inference solely with the SLM. This suggests that effective LLM--SLM collaboration does not necessarily require repeated LLM access, as useful corrective information can be stored and reused for similar reasoning states.

\textbf{Ablation results of two intervention stages.} In Table~\ref{tab:main_results}, the ablation results confirm that prefill- and decode-time interventions are complementary. Removing either component degrades performance, with the reduction being particularly pronounced for the 7B model, where the full RLC achieves 55.42 average accuracy compared with 51.15 and 50.77 for the two single-stage variants. Prefill intervention adjusts the reasoning direction before generating a new step, while decode intervention maintains the corrective signal during token generation. Their combination therefore provides correction at different stages of the reasoning process rather than duplicating the same effect.

\textbf{Intervention behavior across tasks.} Figure~\ref{fig:int-profile} shows that intervention frequency exhibits a clear task-dependent pattern. Competition-level mathematics benchmarks such as AMC24 and AIME require the most frequent intervention, averaging 10.61 and 8.53 interventions, respectively, while GPQA requires only 1.06. MATH500 and GSM8K lie between these two extremes. This pattern suggests that tasks involving longer and more complex reasoning trajectories tend to require more persistent correction, whereas shorter or more localized reasoning processes can often be corrected with fewer interventions. Detailed intervention results are shown in Appendix \ref{app:profile2}.

\vspace{-0.4cm}
\subsection{Cross-Scale Transferability of the Bank (RQ2)}
\begin{figure*}[t]
    \centering
    \includegraphics[width=15.5cm]{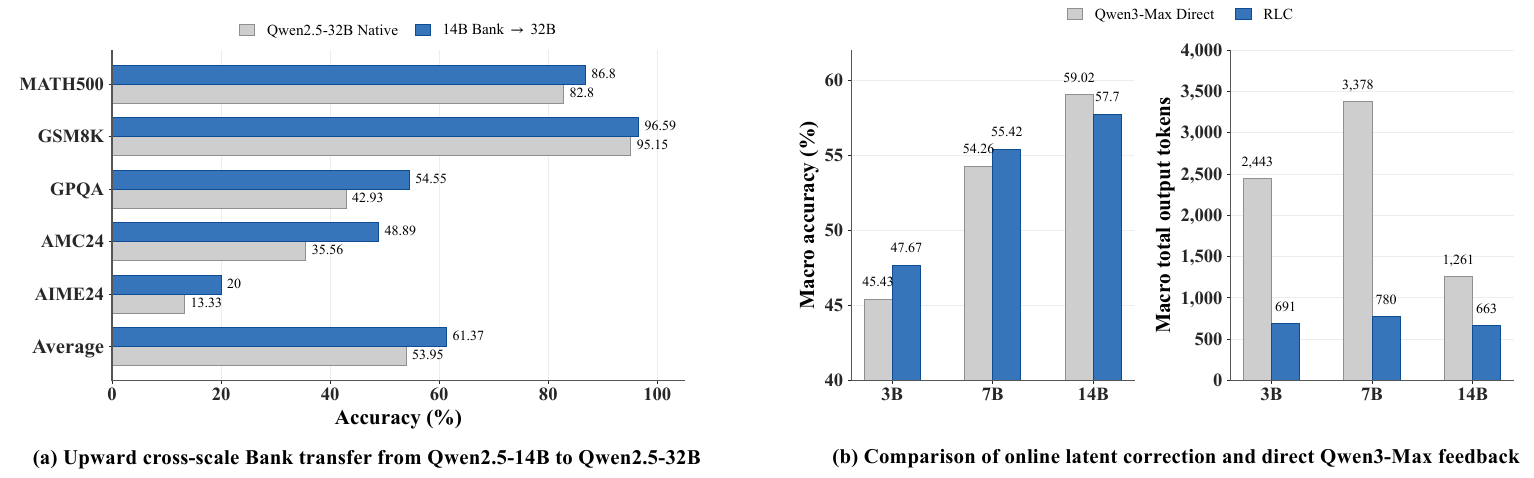}
    \caption{
(a) Upward cross-scale Bank transfer from Qwen2.5-14B to Qwen2.5-32B. The Bank is constructed with the 14B model and applied to the 32B model for inference.
(b) Comparison of online latent correction and direct LLM feedback in terms
of accuracy and total generated output tokens, where the token count for direct feedback includes both local-model and API outputs.
    }
    \label{fig:direct_api}
\end{figure*}

\vspace{-0.2cm}
\label{sec:rq3}
To evaluate whether the Bank can transfer across different model scales, we conduct cross-scale transfer experiments by applying a Bank constructed from one SLM to another SLM of a different scale. Figure~\ref{fig:direct_api}(a) shows the upward-transfer setting, where a Bank built from Qwen2.5-14B is applied to Qwen2.5-32B. As shown in Figure~\ref{fig:direct_api}(a), applying the 14B Bank to Qwen2.5-32B improves the average accuracy by 7.42 percentage points, with consistent gains across all five benchmarks. This result indicates that the correction information in the Bank is not restricted to the model scale from which it is collected. Although the corrections are extracted from the internal states of the 14B model, they can still provide effective guidance to a stronger 32B model, suggesting that the Bank captures transferable reasoning-error patterns rather than model-specific output behavior. We also evaluate the reverse 32B$\rightarrow$14B transfer, with full bidirectional results reported in Appendix~\ref{app:cross_scale_table}.


\subsection{Online Latent Correction vs. Online LLM Feedback (RQ3)}
\label{sec:rq4}
\vspace{-0.2cm}
To evaluate whether offline latent corrections can replace direct LLM feedback during inference, we compare RLC with a direct-feedback variant that queries Qwen3-Max at each retrieved candidate step under the same router and retrieval trigger. Figure~\ref{fig:direct_api}(b) reports the average accuracy and total output tokens. RLC achieves comparable or better accuracy at the 3B and 7B scales while reducing total output tokens by over 70\%. At the 14B scale, its accuracy is only 1.32 percentage points lower, with nearly half the output tokens. These results indicate that repeated online LLM feedback is not necessary for effective correction, as offline guidance can retain much of its effect after being converted into latent corrections. Full per-dataset results are provided in Appendix~\ref{app:direct_table}.



\begin{table*}[t]
\centering
\begingroup
\caption{
Inference efficiency comparison of RLC with same-scale and larger Native models. Results report the average inference time per problem, with lower values indicating better efficiency.
}
\vspace{0.1cm}
\label{tab:inference_time}
\endgroup

\renewcommand{\arraystretch}{0.9}
\setlength{\tabcolsep}{4.4mm}
      \scalebox{0.8}{
\begin{tabular}{llcccccc}
\toprule
Target
& Configuration
& MATH500
& GSM8K
& GPQA
& AMC24
& AIME
& Average \\
\midrule

& 3B Native
& 11.83
& 5.94
& 13.73
& 27.12
& \textbf{16.76}
& \textbf{15.08} \\

3B
& 3B RLC
& \textbf{11.33}
& \textbf{5.44}
& \textbf{12.92}
& 29.10
& 19.70
& 15.70 \\

& 7B Native
& 13.12
& 6.65
& 13.34
& \textbf{24.34}
& 33.09
& 18.11 \\

\midrule

& 7B Native
& \textbf{13.12}
& 6.65
& \textbf{13.34}
& \textbf{24.34}
& 33.09
& \textbf{18.11} \\

7B
& 7B RLC
& 13.49
& \textbf{6.45}
& 15.15
& 26.69
& 31.56
& 18.67 \\

& 32B Native
& 16.73
& 7.34
& 15.60
& 29.26
& \textbf{27.76}
& 19.34 \\

\midrule

& 14B Native
& 18.90
& 8.70
& \textbf{13.59}
& \textbf{19.02}
& 27.68
& \textbf{17.58} \\

14B
& 14B RLC
& \textbf{18.44}
& \textbf{8.42}
& 15.97
& 21.81
& \textbf{26.93}
& 18.31 \\

& 72B Native
& 24.00
& 13.78
& 26.61
& 37.41
& 49.63
& 30.29 \\

\bottomrule
\end{tabular}}
\end{table*}
\subsection{Inference Efficiency of RLC (RQ4)}
\vspace{-0.2cm}
\label{sec:inference_efficiency}
To evaluate the inference overhead introduced by Bank retrieval and latent intervention, we measure the average wall-clock time per problem for RLC, its same-scale Native counterpart, and a larger standalone Native model. Timing covers the full inference process from input preprocessing and tokenization to final decoding. The results are shown in Table~\ref{tab:inference_time}. Compared with same-scale Native inference, RLC introduces only a small latency overhead, averaging 3.1\%--4.2\% across the three model scales. The overhead varies across model--dataset pairs, and RLC is even faster than Native in several cases. This is because retrieval and latent intervention add computation, while successful correction can shorten or simplify subsequent reasoning trajectories. Overall, the additional cost of consulting the Bank remains limited. More importantly, RLC preserves the inference advantage of smaller local models. Across the three comparison groups, RLC is faster than the corresponding larger Native model in 13 of 15 model--dataset comparisons, with a maximum latency reduction of 39.5\% for 14B RLC over 72B Native. This shows that RLC can approach the reasoning performance of much larger models while retaining substantially lower inference latency.

\section{Conclusion}
In this paper, we studied how to improve SLM reasoning while keeping model parameters frozen and avoiding online LLM access. We proposed \textbf{Reusable Latent Correction (RLC)}, which consisted of offline experience construction and online reuse. Offline, RLC identified uncertain reasoning steps, used LLM guidance to regenerate them, and extracted latent corrections from the resulting hidden-state changes. Effective corrections were stored as key--value pairs, linking pre-error reasoning states to correction vectors. Online, the SLM retrieved relevant corrections from its current hidden state and injected them into subsequent generation. Experiments across multiple reasoning benchmarks and model scales demonstrated the effectiveness of RLC.

\bibliography{main}
\newpage
\appendix
\section{Details of Benchmarks}
\label{app:benchmarks}

\subsection{Evaluation Benchmarks}

We evaluate RLC on five reasoning benchmark families covering grade-school arithmetic, general mathematical reasoning, graduate-level scientific question answering, and competition mathematics.

\textbf{MATH500} \citep{lightman2023letsverifystepstep} \footnote{\url{https://huggingface.co/datasets/HuggingFaceH4/MATH-500}} is a 500-problem evaluation subset derived from the MATH benchmark. It covers seven mathematical subjects: algebra, counting and probability, geometry, intermediate algebra, number theory, precalculus, and prealgebra, and includes problems from difficulty levels 1 through 5. We use all examples from the \texttt{test} split. Each problem requires an open-form mathematical answer.

\textbf{GSM8K} \citep{cobbe2021trainingverifierssolvemath} \footnote{\url{https://huggingface.co/datasets/openai/gsm8k}} evaluates multi-step grade-school mathematical reasoning. We use all 1,319 examples from the \texttt{main/test} split. Each reference answer contains a natural-language solution followed by the final numerical answer after the \texttt{\#\#\#\#} delimiter. Only this final numerical answer is used for evaluation.

\textbf{GPQA-Diamond} \citep{rein2024gpqa} \footnote{\url{https://huggingface.co/datasets/Idavidrein/gpqa}} contains graduate-level questions in biology, physics, and chemistry. We use all 198 examples from the \texttt{gpqa\_diamond/train} split. Although this subset is distributed under a split named \texttt{train}, it is used exclusively for evaluation in our experiments. Each question has one correct answer and three incorrect alternatives. Following the standard GPQA evaluation protocol, we shuffle the four alternatives separately for each question using a fixed random seed of 42 and remap the correct option label accordingly.

\textbf{AMC24} \footnote{\url{https://huggingface.co/datasets/rawsh/2024\_AMC12}} is constructed from the 2024 AMC 12A and AMC 12B problems. The data source removes five problems that require figures: Problems 14, 18, and 22 from AMC 12A and Problems 7 and 19 from AMC 12B, leaving 45 text-only problems. We use all 45 remaining problems and compare the final open-form answer with the provided reference answer.

\textbf{AIME24 and AIME25} \footnote{\url{https://huggingface.co/datasets/math-ai/aime24}} \footnote{\url{https://huggingface.co/datasets/math-ai/aime25}} contain the 30 problems from the AIME I and AIME II examinations in 2024 and 2025, respectively. We use the \texttt{test} split of each dataset. These benchmarks require short integer answers and emphasize multi-step competition-level reasoning. The 3B experiments use AIME25, whereas the 7B and 14B experiments use AIME24. This choice is kept fixed across all methods compared in each model~scale.

\subsection{Offline Bank Corpus}
We first screen the full training splits of GSM8K and MATH, comprising 7,473 and 7,500 problems, respectively, for a total of 14,973 examples. For each model scale, the corresponding local model performs a single greedy Native generation on every problem. Problems answered incorrectly in this pass are retained as a model-specific offline corpus, from which we identify candidate reasoning errors and construct reusable latent corrections. Importantly, the Bank is constructed exclusively from these training examples. No downstream evaluation instance from MATH500, the GSM8K test split, GPQA-Diamond, AMC24, AIME24, or AIME25 is used during Bank construction. The problem-level Router is trained on a separate set of examples, as detailed in Appendix~\ref{app:router}.


\section{Experimental Details}
\label{app:experimental_details}

\subsection{Models and Decoding}
We use Qwen2.5-3B-Instruct, Qwen2.5-7B-Instruct, and Qwen2.5-14B-Instruct as the locally deployed SLMs. Qwen3-Max serves as the black-box LLM during offline experience construction and is not accessed during standard RLC inference. All Qwen2.5 backbone parameters remain frozen throughout Bank construction and downstream evaluation. Qwen2.5-7B, Qwen2.5-32B, and Qwen2.5-72B are additionally reported as standalone Native references. They are used only for comparison and do not act as teachers for RLC. Unless otherwise specified, we use greedy decoding, limit each reasoning step to at most 512 generated tokens, and set the maximum total output length for each problem to 4,096 tokens.

\subsection{Comparison Baselines}
\paragraph{Detailed baseline descriptions.} We compare RLC with four representative baselines from two related lines of work: LLM--SLM collaborative inference and inference-time representation intervention. For all baselines, we keep the benchmark splits, answer extraction, and evaluation protocol consistent with RLC.

\textbf{STEER}~\citep{lee2025steer} is a confidence-guided stepwise model-routing method for cost-efficient reasoning. At each reasoning step, the smaller model first generates or evaluates the next step, and token-level logit confidence scores are aggregated into a step-level confidence estimate. A calibrated routing rule then decides whether the current step should remain with the smaller model or be routed to a larger model. In our comparison, STEER represents online LLM--SLM collaboration: for a backbone SLM of scale $A$, STEER can invoke a stronger model of scale $B$ during inference, denoted as $A\rightarrow B$ in Table~\ref{tab:main_results}. Unlike RLC, STEER does not construct a reusable experience bank and still requires access to the larger model at test time.

\textbf{GlimpRouter}~\citep{zeng2026glimprouter} is a training-free stepwise collaborative-inference method that routes reasoning steps by ``glimpsing'' the first token of the next thought. Instead of generating an entire candidate step before deciding whether to switch models, the smaller model only predicts the initial token distribution of the next step. The entropy of this first-token distribution is used as a lightweight estimate of step difficulty: low-entropy steps are handled by the smaller model, whereas high-entropy steps are delegated to the larger model. As with STEER, we report GlimpRouter under the $A\rightarrow B$ setting. This baseline reduces unnecessary large-model calls compared with full-step verification, but it still depends on online access to a larger model during inference.

\textbf{ReBalance}~\citep{li2026rebalance} is a training-free inference-time steering method designed to balance overthinking and underthinking in reasoning models. It uses confidence-related signals to identify reasoning modes and constructs steering directions from hidden-state prototypes associated with different thinking behaviors. During generation, ReBalance dynamically adjusts the strength and direction of the steering vector according to the current confidence state, aiming to suppress redundant reasoning when the model overthinks and encourage further exploration when it underthinks. In our experiments, ReBalance is applied directly to the same frozen SLM backbones without invoking an external LLM. Compared with RLC, ReBalance relies on global behavior-level steering directions rather than retrieving error-specific corrections conditioned on the current local reasoning state.

\textbf{PDS}~\citep{kayan2025pds} is an inference-time representation-intervention method that constructs input-adaptive steering vectors from reasoning prototypes. It first collects activation differences between Chain-of-Thought and neutral prompting conditions, clusters these differences into a set of reasoning prototypes, and then projects the current input representation onto the prototype set to obtain an instance-specific steering vector. This vector is injected into the model hidden states during inference to enhance latent reasoning without changing model weights or prompts. In our comparison, PDS serves as a prototype-based steering baseline. Unlike RLC, which stores key--value corrections derived from concrete local errors and LLM-guided repairs, PDS relies on global prototype-level steering constructed from input-level representation matching and does not explicitly associate a correction with a specific reasoning state.

\subsection{Offline Bank Construction}
During offline construction, we estimate step-level uncertainty using three output-relative Transformer layers, namely $-4$, $-8$, and $-12$. The hidden states from these layers are projected through the model's output head and compared against the final-layer predictive distribution using Jensen--Shannon divergence. For each candidate reasoning step, we consider the final token of the preceding prefix together with all tokens in the current step. If the resulting sequence contains more than 32 positions, we uniformly subsample at most 32 positions using a fixed stride. The uncertainty score is then computed as the mean divergence over the sampled positions and the three intermediate layers. A reasoning step is identified as a candidate error if its uncertainty score exceeds a predefined construction threshold. For each detected candidate error, we provide Qwen3-Max with the original problem, the accepted reasoning prefix, and the candidate step. Qwen3-Max then produces a localized diagnosis of the current error together with a concrete instruction for regenerating only that step. This guidance is appended to the SLM context as a private correction note and is never exposed as part of the visible reasoning trajectory.

For each successfully repaired step, we construct a key from the reasoning state immediately preceding the detected error. The corresponding value concatenates two complementary correction signals: a \emph{process correction}, defined as the difference between the unguided and guidance-conditioned prefix states, and a \emph{result correction}, defined as the difference between the regenerated and original representations of the current step. A candidate key--value pair is added to the Bank only if the complete corrected trajectory ultimately produces the correct Gold answer. Throughout Bank construction, retrieval, and intervention, all representations are extracted from the fourth Transformer layer counted backward from the output side.

\subsection{Online Retrieval and Intervention}
For Bank-routed problems, the SLM generates one visible reasoning step at a time. After each step, its current hidden state is used to retrieve the most similar keys. The default similarity threshold is 0.95, with the selected validation values lying between 0.94 and 0.96. We retrieve the top-$k$ experiences and aggregate their values using similarity-based weights; $k=3$ is used by default and is reduced only when selected on the validation set. All retrieval settings are fixed before test evaluation. The process and result corrections are controlled by $\alpha_p$ and $\alpha_r$, respectively. We use $(\alpha_p,\alpha_r)=(0.5,0.25)$ as the default configuration. A weaker setting of $(0.2,0.05)$ is used for GSM8K and MATH500 to avoid over-correcting trajectories that are often already reliable, whereas $(0.7,0.3)$ is used for AMC24 and AIME, whose longer competition-level trajectories generally require stronger guidance. The process correction is applied to the final prefix token during prefill, and the result correction is applied to each newly decoded token of the next step.

\subsection{Problem-Level Router Training}
\label{app:router}
The problem-level Router is implemented as a binary sequence classifier initialized from Qwen2.5-0.5B-Instruct. It receives only the raw problem text, with a maximum input length of 768 tokens, generated trajectories, retrieved Bank entries, and Gold answers are not used as Router inputs. The supervision is obtained from disagreements between Native and Bank-augmented inference. Problems for which the Bank corrects a Native failure are labeled \emph{rescued} and assigned to the Bank branch, whereas problems for which the Bank changes a correct Native answer into an incorrect one are labeled \emph{destroyed} and assigned to the Native branch. Cases in which both routes are correct or both are incorrect are not used as Router supervision.

Given a problem $x$, the Router predicts the probabilities of the two inference branches as
\begin{equation}
p_\phi(c\mid x)=\operatorname{softmax}\!\left(f_\phi(x)\right)_c,
\qquad c\in\{0,1\},
\label{eq:router_probability}
\end{equation}
where $c=0$ is the \emph{destroyed} class and selects the Native branch, whereas $c=1$ is the \emph{rescued} class and selects the Bank branch.

The training set combines sampled MATH and GSM8K training problems, GPQA training problems that do not overlap with GPQA-Diamond, and independently constructed MetaMath-style problems based on existing training examples. We keep the rescued and destroyed classes as balanced as possible and exclude all downstream evaluation instances. Each Router is fine-tuned with the standard cross-entropy objective:
\begin{equation}
\mathcal{L}_{\mathrm{router}}
=-\frac{1}{N}\sum_{i=1}^{N}\sum_{c\in\{0,1\}}
\mathbb{I}(y_i=c)\log p_\phi(c\mid x_i).
\label{eq:router_loss}
\end{equation}
At inference time, the Router is called once on the input problem and uses a fixed threshold~$\tau_R=0.5$:
\begin{equation}
r(x)=\mathbb{I}\!\left[p_\phi(y=1\mid x)\geq\tau_R\right].
\label{eq:router_decision}
\end{equation}
When $r(x)=1$, RLC activates Bank retrieval and hidden-state intervention for the subsequent reasoning trajectory; otherwise, the frozen local model completes Native inference.

\subsection{Evaluation}
\label{app:answer_evaluation}
We report Gold-answer accuracy using the official reference answers provided by each benchmark. The evaluator first extracts the final answer from each model output. For MATH500, GSM8K, AMC24, AIME24, and AIME25, the answer contained in the final \verb|\boxed{}| expression is extracted when available. For GSM8K reference answers, the final numerical value following the \texttt{\#\#\#\#} delimiter is used. The extracted prediction and reference answer are then normalized and compared automatically. The normalization procedure removes commas, surrounding whitespace, units, and common formatting wrappers. A prediction is considered correct if it exactly matches the reference answer, is numerically equivalent to it, or corresponds to a symbolically equivalent mathematical expression. For GPQA-Diamond, the evaluator extracts the final predicted option letter and compares it directly with the remapped Gold option letter.

\begin{table*}[t]
\centering
\begingroup
\caption{Effect of state-conditioned correction retrieval.  Mean Value replaces each retrieved correction with the global mean of all correction values in the Bank, removing the correspondence between the current reasoning state and its retrieved correction. }
\vspace{0.1cm}
\label{tab:value_ablation}
\endgroup

\renewcommand{\arraystretch}{0.86}
\setlength{\tabcolsep}{2.3pt}

\resizebox{\textwidth}{!}{%
\begin{tabular}{ll*{5}{cc}cc}
\toprule
Model & Variant
& \multicolumn{2}{c}{MATH500}
& \multicolumn{2}{c}{GSM8K}
& \multicolumn{2}{c}{GPQA}
& \multicolumn{2}{c}{AMC24}
& \multicolumn{2}{c}{AIME$^{\dagger}$}
& \multicolumn{2}{c}{Average} \\
\cmidrule(lr){3-4}
\cmidrule(lr){5-6}
\cmidrule(lr){7-8}
\cmidrule(lr){9-10}
\cmidrule(lr){11-12}
\cmidrule(lr){13-14}
& & Acc. & Tok.
& Acc. & Tok.
& Acc. & Tok.
& Acc. & Tok.
& Acc. & Tok.
& Acc. & Tok. \\
\midrule

3B & \textbf{RLC}
& \textbf{73.00} & 576.91
& \textbf{89.99} & 272.68
& \textbf{40.91} & \textbf{690.92}
& \textbf{24.44} & \textbf{992.78}
& \textbf{10.00} & \textbf{921.93}
& \textbf{47.67} & \textbf{691.04} \\

& Mean Value
& 70.00 & \textbf{574.73}
& 86.28 & \textbf{272.29}
& 36.36 & 723.97
& 20.00 & 1003.78
& \textbf{10.00} & 963.77
& 44.53 & 707.71 \\

\midrule

7B & \textbf{RLC}
& \textbf{79.80} & \textbf{564.70}
& \textbf{94.09} & \textbf{279.07}
& \textbf{45.45} & 654.44
& \textbf{37.78} & 1100.38
& \textbf{20.00} & \textbf{1300.53}
& \textbf{55.42} & 779.82 \\

& Mean Value
& 76.80 & 568.91
& 93.71 & 279.48
& 40.40 & \textbf{652.57}
& 35.56 & \textbf{906.07}
& 10.00 & 1335.27
& 51.29 & \textbf{748.46} \\

\midrule

14B & \textbf{RLC}
& \textbf{83.00} & \textbf{583.87}
& \textbf{95.68} & 280.45
& \textbf{52.02} & \textbf{467.12}
& \textbf{37.78} & \textbf{864.58}
& \textbf{20.00} & 1120.07
& \textbf{57.70} & \textbf{663.22} \\

& Mean Value
& 78.60 & 592.06
& 95.30 & \textbf{280.34}
& 48.51 & 470.31
& 33.33 & 941.69
& 10.00 & \textbf{1105.83}
& 53.15 & 678.05 \\

\bottomrule
\end{tabular}%
}

\vspace{0.5mm}
\begin{minipage}{0.98\textwidth}
\footnotesize
\end{minipage}
\end{table*}

\section{More Experiments}
\subsection{Hyperparameter Sensitivity Analysis}

\subsubsection{Effectiveness of State-conditioned Correction Retrieval}
\label{sec:rq5}
To evaluate whether the effectiveness of RLC depends on matching the current reasoning state with an appropriate correction, we compare the full RLC with a \textit{Mean Value} variant. Mean Value keeps the same Router and retrieval schedule, but replaces the retrieved correction with the global mean of all Bank values, thereby removing the instance-specific correspondence between keys and corrections. As shown in Table~\ref{tab:value_ablation}, replacing the retrieved correction with the global mean consistently degrades performance. Across different model scales, the average accuracy drops by 3.14--4.55 percentage points, and RLC outperforms Mean Value in 14 of the 15 model--dataset comparisons. Since the Router and retrieval positions remain unchanged, this gap mainly reflects the loss of state-specific correction rather than differences in when intervention is applied. This indicates that the benefit of RLC does not come from simply injecting a generic steering direction. Instead, the retrieved correction needs to correspond to the current reasoning state.

\begin{table}[h]
\centering

\begingroup
\caption{
Sensitivity of RLC to the injection layer on Qwen2.5-3B. 
}
\vspace{0.1cm}
\label{tab:injection_layer}
\endgroup

\renewcommand{\arraystretch}{0.95}
\setlength{\tabcolsep}{12pt}

\begin{tabular}{cc}
\toprule
Injection layer & Acc. \\
\midrule
$-2$  & 73.00 \\
$-4$  & \textbf{75.00} \\
$-8$  & 74.00 \\
$-12$ & 74.00 \\
$-18$ & 73.00 \\
\bottomrule
\end{tabular}

\end{table}

\subsubsection{Sensitivity to the Injection Layer}
\label{sec:injection_layer}
To evaluate the sensitivity of RLC to the injection layer, we conduct experiments on Qwen2.5-3B using five output-relative Transformer layers: $\{ -2,-4,-8,-12,-18\}$. For all candidate layers, the corresponding Banks are constructed from the same reasoning trajectories and teacher feedback, differing only in the hidden states used to form the keys and values. During evaluation, each Bank is applied to its corresponding layer, while the Router, retrieval trigger, top-$K$ setting, and all other inference configurations remain unchanged. For each trial, we randomly sample 70 problems from the MATH training split and 30 problems from the GSM8K training split, forming a 100-example in-domain development set.

The results in Table \ref{tab:injection_layer} show that RLC is relatively insensitive to the exact injection layer. The default $-4$th layer achieves the highest mean accuracy of 75.00\%, while the gap between the best and worst candidate layers is only 2.00 percentage points. Since all layer-specific Banks are constructed from identical trajectories and teacher feedback, the observed differences can be attributed primarily to the injection location rather than changes in correction content. This indicates that the stored corrections remain effective across a range of intermediate and late Transformer layers. We further observe that intermediate layers generally perform slightly better than either very early or output-proximal layers. A possible explanation is that earlier layers may provide less abstract reasoning representations, whereas layers close to the output leave less computation for the model to integrate the injected correction. Intermediate layers therefore offer a better balance between representational abstraction and downstream propagation. Based on these results, we use the $-4$th layer as the default injection layer in the remaining experiments.

\subsection{Full Cross-Scale Transfer Results}
\label{app:cross_scale_table}
To evaluate whether the experience Bank can transfer across different model scales, we conduct cross-scale transfer experiments by applying a Bank constructed from one SLM to another SLM of a different scale. Table~\ref{tab:cross_scale_transfer_full} reports the bidirectional cross-scale transfer results between Qwen2.5-14B and Qwen2.5-32B. In each direction, the experience Bank is constructed from one model during the offline stage and then applied to the other model for online inference.

As shown in Table~\ref{tab:cross_scale_transfer_full}, transferring the Bank constructed with Qwen2.5-14B to Qwen2.5-32B improves the average accuracy by 7.42 percentage points, with gains observed on all five benchmarks. This result provides additional evidence that the latent corrections stored in the Bank can generalize beyond the model used for Bank construction. Despite being derived from the internal representations of Qwen2.5-14B, these corrections remain effective when applied to Qwen2.5-32B, demonstrating a certain degree of cross-scale transferability of the learned corrective experience. Besides, we further observe a clear asymmetry in cross-scale transfer. While the 14B Bank provides a 7.42-point improvement to the 32B model, transferring the 32B Bank back to the 14B model yields only a 1.08-point gain. One possible reason is that smaller models expose more frequent and diverse reasoning failures during Bank construction, resulting in broader coverage of reusable error patterns. In contrast, the errors of larger models are fewer and more concentrated on difficult cases, making the resulting Bank less comprehensive. Meanwhile, a stronger receiving model has greater capacity to follow a transferred correction and continue the subsequent reasoning process, whereas a smaller model may still fail even when the local reasoning direction is corrected. These results suggest that cross-scale transfer is more effective from a smaller-model Bank to a stronger receiver.

\begin{table*}[!t]
\centering
\caption{
Full cross-scale Bank transfer results.
}
\vspace{0.1cm}
\label{tab:cross_scale_transfer_full}
\renewcommand{\arraystretch}{0.90}
\setlength{\tabcolsep}{2.2pt}
\resizebox{\textwidth}{!}{%
\begin{tabular}{l*{5}{ccc}cc}
\toprule
\multirow{2}{*}{Method}
& \multicolumn{3}{c}{MATH500}
& \multicolumn{3}{c}{GSM8K}
& \multicolumn{3}{c}{GPQA}
& \multicolumn{3}{c}{AMC24}
& \multicolumn{3}{c}{AIME24}
& \multicolumn{2}{c}{Average} \\
\cmidrule(lr){2-4}
\cmidrule(lr){5-7}
\cmidrule(lr){8-10}
\cmidrule(lr){11-13}
\cmidrule(lr){14-16}
\cmidrule(lr){17-18}
& Acc. & Tok. & Int.
& Acc. & Tok. & Int.
& Acc. & Tok. & Int.
& Acc. & Tok. & Int.
& Acc. & Tok. & Int.
& Acc. & Tok. \\
\midrule
\multicolumn{18}{l}{\textit{32B Bank $\rightarrow$ 14B}} \\
Qwen2.5-14B Native
& 76.20 & 602.48 & --
& 94.47 & 290.74 & --
& 42.93 & 634.18 & --
& 31.11 & 766.82 & --
& 13.33 & 1122.87 & --
& 51.61 & 683.42 \\
32B Bank $\rightarrow$ 14B
& \textbf{78.40} & \textbf{561.40} & 3.16
& \textbf{95.45} & \textbf{280.93} & 0.37
& 42.93 & 647.59 & 0.32
& \textbf{33.33} & 1138.64 & 11.74
& 13.33 & 1363.53 & 16.30
& \textbf{52.69} & 798.42 \\
\midrule
\multicolumn{18}{l}{\textit{14B Bank $\rightarrow$ 32B}} \\
Qwen2.5-32B Native
& 82.80 & 530.40 & --
& 95.15 & 231.24 & --
& 42.93 & \textbf{481.80} & --
& 35.56 & 947.33 & --
& 13.33 & \textbf{839.73} & --
& 53.95 & \textbf{606.10} \\
14B Bank $\rightarrow$ 32B
& \textbf{86.80} & \textbf{507.06} & 3.03
& \textbf{96.59} & \textbf{229.13} & 2.13
& \textbf{54.55} & 512.77 & 0.70
& \textbf{48.89} & \textbf{908.02} & 4.20
& \textbf{20.00} & 960.67 & 3.39
& \textbf{61.37} & 623.53 \\
\bottomrule
\end{tabular}}
\end{table*}

\begin{table*}[t]
\centering
\begingroup
\caption{
Effectiveness of the Router across different model scales and reasoning benchmarks. 
}
\vspace{0.1cm}
\label{tab:router-effectiveness}
\endgroup

\renewcommand{\arraystretch}{0.90}
\setlength{\tabcolsep}{3.0pt}

\resizebox{\textwidth}{!}{%
\begin{tabular}{l*{5}{cc}c}
\toprule
\multirow{2}{*}{Route}
& \multicolumn{2}{c}{{MATH500}}
& \multicolumn{2}{c}{{GSM8K}}
& \multicolumn{2}{c}{{GPQA}}
& \multicolumn{2}{c}{{AMC24}}
& \multicolumn{2}{c}{{AIME25}}
& \multirow{2}{*}{{Avg. Acc.}} \\
\cmidrule(lr){2-3}
\cmidrule(lr){4-5}
\cmidrule(lr){6-7}
\cmidrule(lr){8-9}
\cmidrule(lr){10-11}
& Acc. & Bank (\%)
& Acc. & Bank (\%)
& Acc. & Bank (\%)
& Acc. & Bank (\%)
& Acc. & Bank (\%)
& \\
\midrule
\multicolumn{12}{l}{\small\textit{Qwen2.5-3B-Instruct}} \\
Native
& 67.00 & 0.00
& 84.69 & 0.00
& 30.30 & 0.00
& 4.44  & 0.00
& 3.33  & 0.00
& 37.95 \\
Bank
& 65.00 & 100.00
& 83.70 & 100.00
& 31.82 & 100.00
& 22.22 & 100.00
& 6.67  & 100.00
& 41.88 \\
\textbf{Router}
& \textbf{73.00} & 50.60
& \textbf{89.99} & 40.71
& \textbf{40.91} & 48.99
& \textbf{24.44} & 60.00
& \textbf{10.00} & 73.33
& \textbf{47.67} \\
\bottomrule
\end{tabular}%
}

\vspace{1.0mm}

\resizebox{\textwidth}{!}{%
\begin{tabular}{l*{5}{cc}c}
\toprule
\multirow{2}{*}{Route}
& \multicolumn{2}{c}{{MATH500}}
& \multicolumn{2}{c}{{GSM8K}}
& \multicolumn{2}{c}{{GPQA}}
& \multicolumn{2}{c}{{AMC24}}
& \multicolumn{2}{c}{{AIME24}}
& \multirow{2}{*}{{Avg. Acc.}} \\
\cmidrule(lr){2-3}
\cmidrule(lr){4-5}
\cmidrule(lr){6-7}
\cmidrule(lr){8-9}
\cmidrule(lr){10-11}
& Acc. & Bank (\%)
& Acc. & Bank (\%)
& Acc. & Bank (\%)
& Acc. & Bank (\%)
& Acc. & Bank (\%)
& \\
\midrule
\multicolumn{12}{l}{\small\textit{Qwen2.5-7B-Instruct}} \\
Native
& 74.20 & 0.00
& 90.83 & 0.00
& 38.89 & 0.00
& 20.00 & 0.00
& 13.33 & 0.00
& 47.45 \\
Bank
& 74.00 & 100.00
& 90.75 & 100.00
& 37.88 & 100.00
& 35.56 & 100.00
& 16.67 & 100.00
& 50.97 \\
\textbf{Router}
& \textbf{79.80} & 50.40
& \textbf{94.09} & 32.30
& \textbf{45.45} & 58.08
& \textbf{37.78} & 71.11
& \textbf{20.00} & 76.67
& \textbf{55.42} \\
\bottomrule
\end{tabular}%
}

\vspace{1.0mm}

\resizebox{\textwidth}{!}{%
\begin{tabular}{l*{5}{cc}c}
\toprule
\multirow{2}{*}{Route}
& \multicolumn{2}{c}{{MATH500}}
& \multicolumn{2}{c}{{GSM8K}}
& \multicolumn{2}{c}{{GPQA}}
& \multicolumn{2}{c}{{AMC24}}
& \multicolumn{2}{c}{{AIME24}}
& \multirow{2}{*}{{Avg. Acc.}} \\
\cmidrule(lr){2-3}
\cmidrule(lr){4-5}
\cmidrule(lr){6-7}
\cmidrule(lr){8-9}
\cmidrule(lr){10-11}
& Acc. & Bank (\%)
& Acc. & Bank (\%)
& Acc. & Bank (\%)
& Acc. & Bank (\%)
& Acc. & Bank (\%)
& \\
\midrule
\multicolumn{12}{l}{\small\textit{Qwen2.5-14B-Instruct}} \\
Native
& 77.00 & 0.00
& 94.47 & 0.00
& 42.93 & 0.00
& 31.11 & 0.00
& 13.33 & 0.00
& 51.77 \\
Bank
& 79.40 & 100.00
& 93.25 & 100.00
& 41.41 & 100.00
& \textbf{37.78} & 100.00
& 16.67 & 100.00
& 53.70 \\
\textbf{Router}
& \textbf{83.00} & 50.00
& \textbf{95.68} & 14.48
& \textbf{52.02} & 42.42
& \textbf{37.78} & 77.78
& \textbf{20.00} & 76.67
& \textbf{57.70} \\
\bottomrule
\end{tabular}%
}

\end{table*}

\subsection{Effectiveness of the Router}
\label{sec:rq2}
To evaluate whether the Router can effectively determine when Bank-based correction is needed, we compare three inference strategies: \textit{Native}, which always uses the original SLM without accessing the Bank. \textit{Bank}, which always activates Bank-augmented inference, and \textit{Router}, which selectively chooses between the two branches. The results are shown in Table~\ref{tab:router-effectiveness}. The Router achieves the best accuracy in all 15 model--dataset comparisons, outperforming the better of the two fixed strategies by at least 4.00 percentage points in average accuracy at each model scale. More importantly, the advantage of the Router over always using the Bank reveals that Bank augmentation is not a uniformly beneficial operation. Although the stored corrections are constructed from successful error correction, their effectiveness remains state-dependent at test time: a retrieved correction may be useful when the current reasoning state exhibits a similar error pattern, but unnecessary or mismatched intervention can perturb an otherwise correct trajectory. Therefore, routing every problem through the Bank inevitably applies corrections to instances for which no correction is needed.

Besides, a second observation is that the Router becomes increasingly selective as model capability changes, rather than simply reducing Bank usage uniformly. For example, the Bank rate on GSM8K decreases from 40.71\% for the 3B model to only 14.48\% for the 14B model, while it remains high on AMC24 and AIME across model scales. This suggests that the Router responds to the residual reasoning difficulty of each model--task combination: stronger SLMs rely less on external correction for problems they can already handle reliably, but still frequently activate the Bank on challenging problems where substantial reasoning errors remain.

\subsection{Per-Dataset Direct-Intervention Results}
\label{app:direct_table}

Table~\ref{tab:direct_intervention_full} provides the per-dataset values summarized by Figure~\ref{fig:direct_api}(b).

\subsection{Detailed Intervention Statistics}
\label{app:profile2}
Table~\ref{tab:int-profile2} reports the full intervention statistics under Bank routing across the 3B, 7B, and 14B models on all five benchmarks. Beyond the observations discussed in Section \ref{sec:rq1}, we also find the intervention frequency does not decrease monotonically as the SLM becomes larger. This indicates that the number of interventions is not determined solely by model capacity.

\begin{table*}[!t]
\centering
\caption{
Comparison between online latent correction and direct LLM feedback using Qwen3-Max. Native is omitted because it does not involve any intervention. A dash indicates that RLC makes no online API calls. \textit{Small} denotes the number of output tokens generated by the SLM, \textit{API} denotes the number of output tokens generated by Qwen3-Max, and \textit{Total} is the sum of the two.
}
\vspace{0.1cm}
\label{tab:direct_intervention_full}
\renewcommand{\arraystretch}{1.50}
\setlength{\tabcolsep}{2.8pt}
\resizebox{\textwidth}{!}{%
\begin{tabular}{ll*{5}{rrrr}}
\toprule
\multirow{2}{*}{Model}
& \multirow{2}{*}{Method}
& \multicolumn{4}{c}{MATH500}
& \multicolumn{4}{c}{GSM8K}
& \multicolumn{4}{c}{GPQA}
& \multicolumn{4}{c}{AMC24}
& \multicolumn{4}{c}{AIME} \\
\cmidrule(lr){3-6}
\cmidrule(lr){7-10}
\cmidrule(lr){11-14}
\cmidrule(lr){15-18}
\cmidrule(lr){19-22}
& & Acc. & Small & API & Total
& Acc. & Small & API & Total
& Acc. & Small & API & Total
& Acc. & Small & API & Total
& Acc. & Small & API & Total \\
\midrule
3B & Qwen3-Max Direct
& 70.20 & 792.68 & 425.23 & 1217.91
& 87.82 & 297.15 & 126.13 & 423.28
& 36.90 & 896.59 & 308.09 & 1204.68
& 22.22 & 1529.93 & 1312.67 & 2842.60
& 10.00 & 2493.67 & 4033.10 & 6526.77 \\
& RLC
& 73.00 & 576.91 & -- & 576.91
& 89.99 & 272.68 & -- & 272.68
& 40.91 & 690.92 & -- & 690.92
& 24.44 & 992.78 & -- & 992.78
& 10.00 & 921.93 & -- & 921.93 \\
\midrule
7B & Qwen3-Max Direct
& 79.60 & 757.36 & 556.52 & 1313.88
& 92.49 & 302.68 & 76.49 & 379.17
& 46.97 & 643.12 & 52.03 & 695.15
& 35.56 & 2329.78 & 3437.78 & 5767.56
& 16.67 & 2618.57 & 6114.10 & 8732.67 \\
& RLC
& 79.80 & 564.70 & -- & 564.70
& 94.09 & 279.07 & -- & 279.07
& 45.45 & 654.44 & -- & 654.44
& 37.78 & 1100.38 & -- & 1100.38
& 20.00 & 1300.53 & -- & 1300.53 \\
\midrule
14B & Qwen3-Max Direct
& 81.60 & 668.33 & 452.76 & 1121.09
& 94.92 & 292.36 & 61.42 & 353.78
& 53.03 & 481.49 & 77.26 & 558.76
& 42.22 & 934.42 & 994.51 & 1928.93
& 23.33 & 1241.53 & 1100.77 & 2342.30 \\
& RLC
& 83.00 & 583.87 & -- & 583.87
& 95.68 & 280.45 & -- & 280.45
& 52.02 & 467.12 & -- & 467.12
& 37.78 & 864.58 & -- & 864.58
& 20.00 & 1120.07 & -- & 1120.07 \\
\bottomrule
\end{tabular}}
\end{table*}

\begin{table*}[!t]
\caption{
Intervention behavior across tasks.
We report the average number of interventions among Bank-routed problems.
The rightmost column reports the mean across the five datasets for each model,
whereas the bottom row reports the mean across the RLC 3B, 7B, and 14B for
each dataset. 
}
\vspace{0.1cm}
\label{tab:int-profile2}
\centering
\setlength{\tabcolsep}{8.0pt}
\renewcommand{\arraystretch}{1.08}
\resizebox{0.8\textwidth}{!}{%
\begin{tabular}{lcccccc}
\toprule
Model
& MATH500
& GSM8K
& GPQA
& AMC24
& AIME
& Average \\
\midrule
RLC (3B)
& 5.09
& 3.25
& 1.80
& 12.52
& 10.73
& \textbf{6.68} \\

RLC (7B)
& 4.27
& 1.69
& 0.35
& 8.84
& 7.13
& \textbf{4.46} \\

RLC (14B)
& 9.15
& 9.00
& 1.04
& 10.46
& 7.74
& \textbf{7.48} \\
\midrule
\textbf{Average}
& \textbf{6.17}
& \textbf{4.65}
& \textbf{1.06}
& \textbf{10.61}
& \textbf{8.53}
& \textbf{6.20} \\
\bottomrule
\end{tabular}%
}
\end{table*}

\subsection{Case Studies}
\label{app:case_studies}

\begin{figure*}[t]
\centering
\includegraphics[width=16.cm]{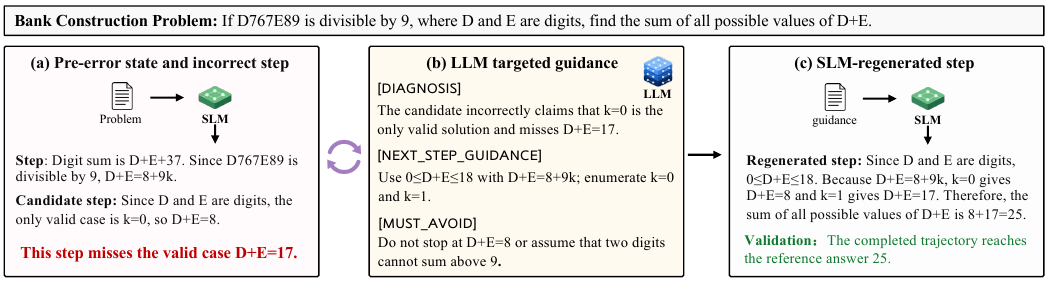}
\caption{
Case study of offline Bank construction in RLC. An incorrect reasoning step is identified, corrected with localized LLM guidance, and converted into a validated reusable latent experience.
}
\label{fig:case_bank_construction}
\end{figure*}

\begin{figure*}[t]
\centering
\includegraphics[width=16cm]{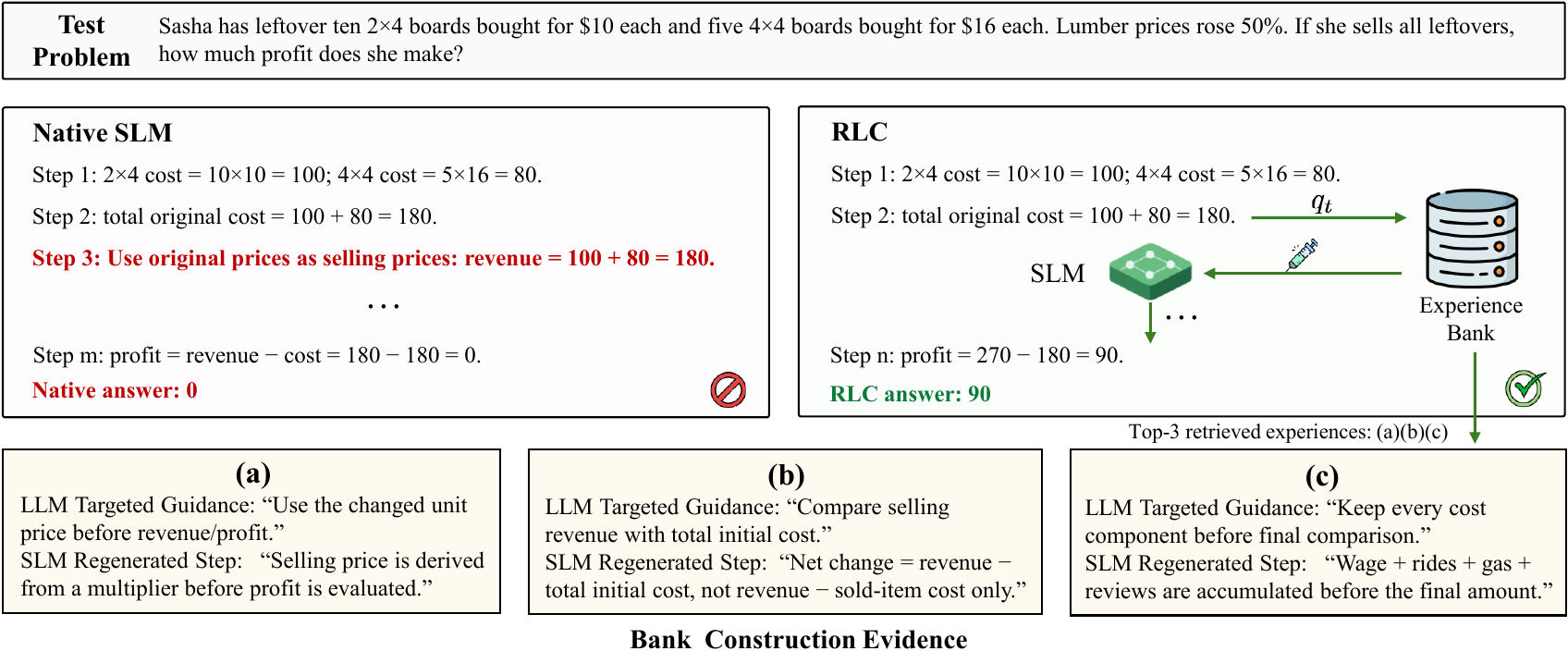}
\caption{%
Case study of online latent experience reuse on GSM8K. RLC retrieves relevant latent Bank entries to correct a local reasoning error and steer the frozen SLM toward the correct solution. The displayed Bank Construction Evidence is provided only to interpret the retrieved latent experiences and is not used during online inference.
}
\vspace{-0.3cm}
\label{fig:case_gsm8k}
\end{figure*}

\begin{figure*}[t]
\centering
\includegraphics[width=16cm]{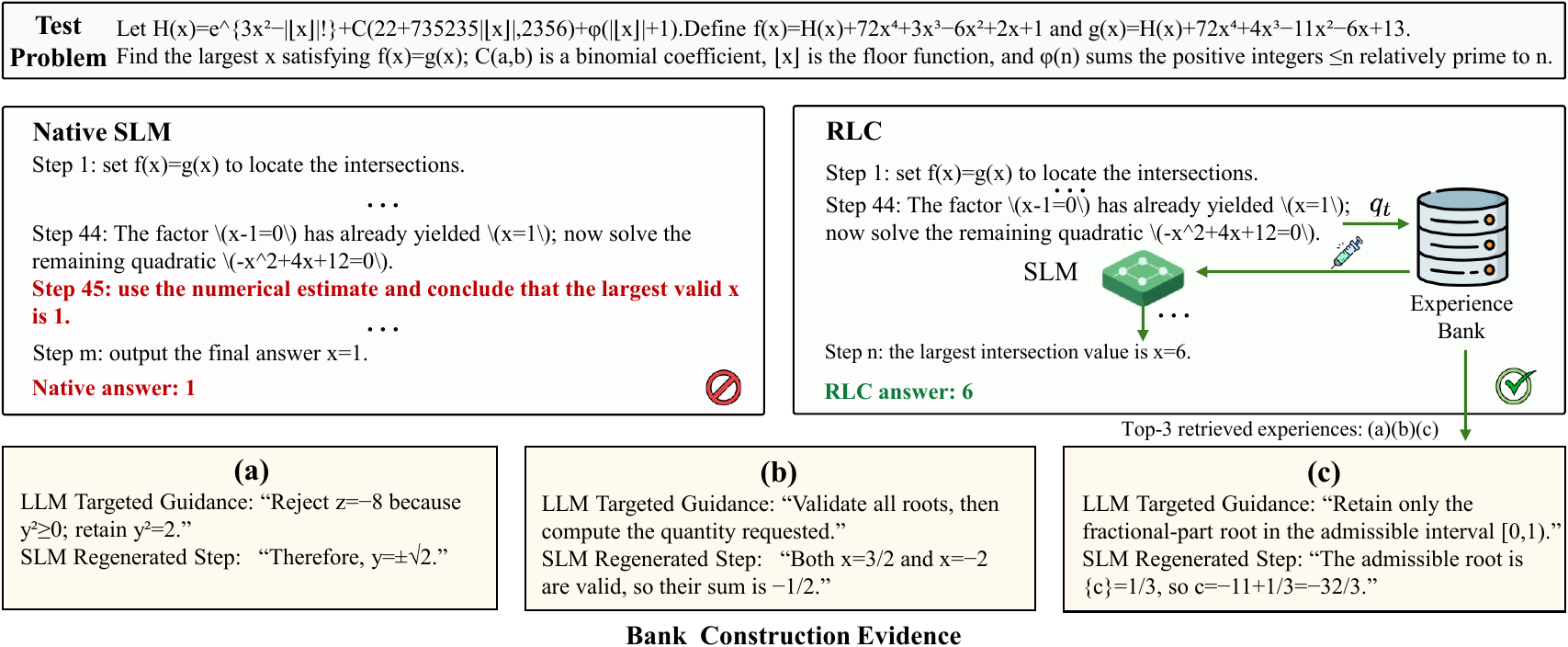}
\caption{%
Case study of online latent experience reuse on MATH500. RLC retrieves relevant latent Bank entries to correct a local reasoning error and steer the frozen SLM toward the correct solution. The displayed Bank Construction Evidence is provided only to interpret the retrieved latent experiences and is not used during online inference.
}
\label{fig:case_math500}
\end{figure*}

\begin{figure*}[t]
\centering
\includegraphics[width=16cm]{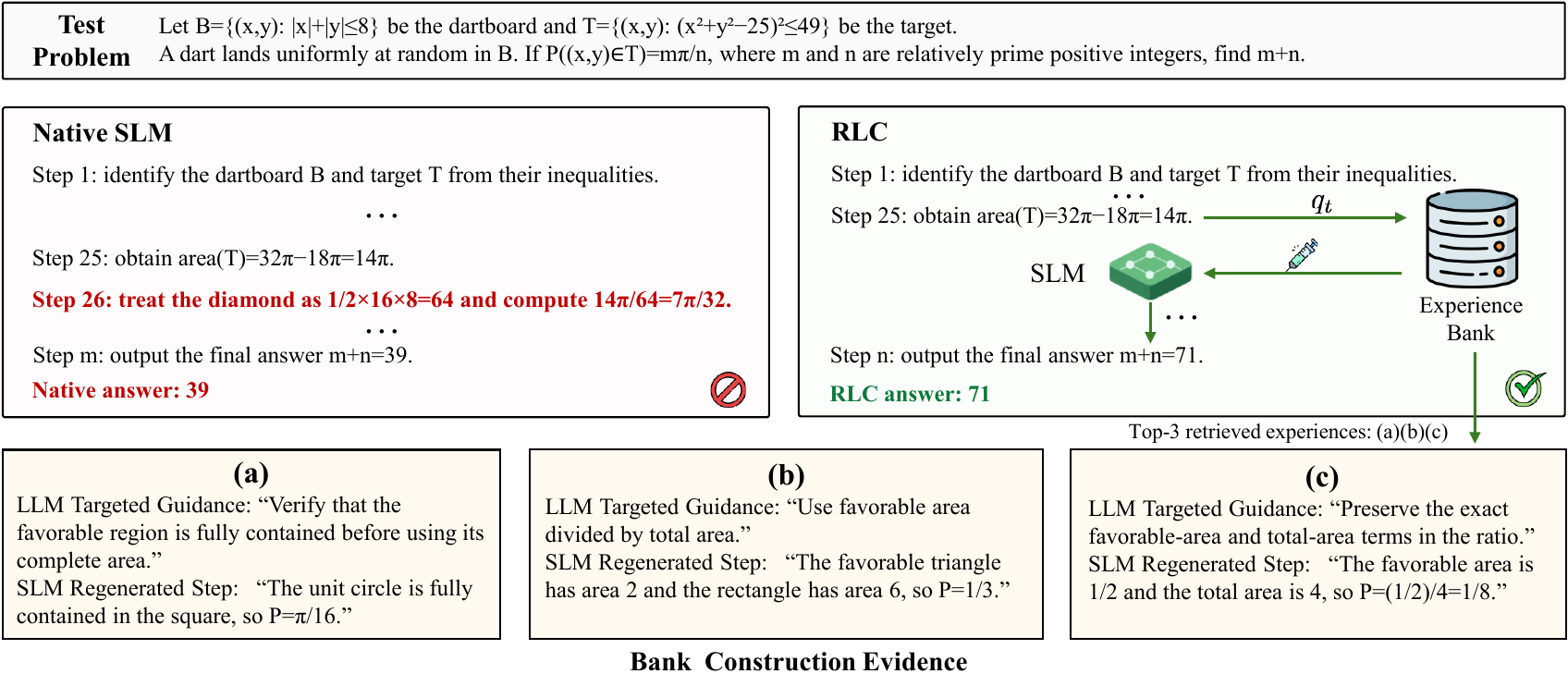}
\caption{%
Case study of online latent experience reuse on AMC24. RLC retrieves relevant latent Bank entries to correct a local reasoning error and steer the frozen SLM toward the correct solution. The displayed Bank Construction Evidence is provided only to interpret the retrieved latent experiences and is not used during online inference.
}
\label{fig:case_amc24}
\end{figure*}

We present qualitative case studies to illustrate the two stages of RLC: offline experience construction and online experience reuse. The first case (Figure~\ref{fig:case_bank_construction}) shows how a localized teacher repair is converted into a reusable key--value latent experience. The remaining three cases (Figures~\ref{fig:case_gsm8k}, \ref{fig:case_math500}, and \ref{fig:case_amc24}) show how retrieved latent corrections modify test-time reasoning trajectories across GSM8K, MATH500, and AMC24. These examples make the mechanism of RLC concrete: the Bank does not store natural-language feedback, and RLC does not call the LLM during online inference. Instead, it reuses validated hidden-state corrections to steer a frozen SLM at local reasoning states.

\textbf{Offline Bank construction.} Figure~\ref{fig:case_bank_construction} illustrates how RLC constructs a Bank entry from a local repair. As shown in Figure~\ref{fig:case_bank_construction}(a), the Native trajectory first reaches a pre-error reasoning state and then produces an incorrect candidate step $s_t^-$ (e.g., ``Since D and E are digits, the only valid case is $k=0$, so $D+E=8$.''). In Figure~\ref{fig:case_bank_construction}(b), the black-box LLM provides targeted guidance for this uncertain step, consisting of three components: \texttt{[DIAGNOSIS]}, \texttt{[NEXT\_STEP\_GUIDANCE]}, and \texttt{[MUST\_AVOID]}. Conditioned on this guidance, the frozen SLM regenerates the corresponding corrected step $s_t^+$, as shown in Figure~\ref{fig:case_bank_construction}(c). The regenerated trajectory is retained only if its final answer is verified as correct, so that the resulting Bank entry is constructed from a validated local correction. RLC stores the effect of this repair in hidden-state space: the key represents the reasoning state immediately before the wrong step, while the value contains the process and result correction components induced by the guided regeneration. Thus, the stored experience is not the teacher text itself, but a reusable latent correction associated with a concrete pre-error state.

\noindent\textbf{Online experience reuse.} Figure~\ref{fig:case_gsm8k} presents a GSM8K example illustrating how RLC corrects a local reasoning error through latent experience reuse. The Native SLM correctly computes the total original cost as \$180, but makes an error when estimating the selling revenue: it directly reuses the original prices and obtains a revenue of \$180, ignoring the 50\% price increase. This local mistake propagates to the final answer, yielding an incorrect profit of zero. At this reasoning state, RLC queries the Bank and retrieves the top-3 relevant latent experiences. The retrieved entries are represented by latent keys and values, rather than natural-language instructions. After their latent values are aggregated and injected into the frozen SLM, the subsequent reasoning trajectory is corrected: the model accounts for the increased selling price, obtains a revenue of \$270, and finally produces the correct profit of \$90. Since the retrieved Bank entries are latent vectors and cannot be directly visualized, Figure~\ref{fig:case_gsm8k} additionally presents their corresponding \emph{Bank Construction Evidence}. For each retrieved latent experience, we show the LLM-provided targeted guidance and the SLM-regenerated reasoning step that were used during offline construction of that entry. These textual elements are shown only to make the retrieved latent experiences interpretable; they are neither stored as the online retrieval content nor reused during test-time inference. Figures~\ref{fig:case_math500} and~\ref{fig:case_amc24} provide two additional examples exhibiting similar local correction behavior on other reasoning tasks.

\vspace{-0.3cm}
\subsection{Bank Construction Cost and Amortized Reuse}
\vspace{-0.2cm}
We further analyze the monetary cost of constructing and reusing the RLC Bank. All costs reported in this section refer only to Qwen3-Max API usage and exclude local SLM computation. Based on the Qwen3-Max pricing at the time of our experiments\footnote{\url{https://docs.modelstudio.console.alibabacloud.com/zh/model-studio/model-pricing}}, we convert the recorded input and output token usage into U.S. dollars using an exchange rate of \(1~\mathrm{USD}=6.6977~\mathrm{CNY}\).

\noindent\textbf{Offline Bank construction.} Averaged across the Qwen2.5-3B, 7B, and 14B construction processes, each problem requires \(8.43\) Qwen3-Max calls and \(1{,}448.58\) API output tokens on average. Constructing all three Banks incurs a total teacher-API cost of approximately \(\mathrm{US}\$52.80\).

\noindent\textbf{Repeated online feedback.} For comparison, in the direct-feedback evaluation reported in Table~\ref{tab:direct_intervention_full}, each problem receiving direct feedback requires \(4.65\) Qwen3-Max calls and \(738.52\) API output tokens on average. A complete direct-feedback evaluation sweep over the three model scales and five benchmarks costs approximately \(\mathrm{US}\$7.10\).

\noindent\textbf{Amortized reuse.} The above comparison highlights an important difference between the two paradigms. Direct feedback incurs additional API cost for every new problem, whereas the constructed Bank can be reused without further teacher-API calls or SLM parameter updates. The one-time Bank construction cost of \(\mathrm{US}\$52.80\) is equivalent to approximately \(52.80/7.10=7.44\) complete direct-feedback evaluation sweeps of the scale reported in Table~\ref{tab:direct_intervention_full}. As the Bank is reused across more benchmarks, users, and queries, its amortized teacher-API cost per downstream instance continually decreases.

\clearpage
\onecolumn

\section{Model Prompts}
\label{app:model_prompts}
\vspace{-0.5em}
\begin{tcolorbox}[
enhanced,
breakable,
before skip=0pt,
colback=gray!3,
colframe=SEUGreenDark,
colbacktitle=SEUGreenDark,
coltitle=white,
title={Stepwise generation instructions used for Bank trajectories},
fonttitle=\bfseries,
boxrule=0.6pt,
arc=1mm,
width=\textwidth,
boxsep=1mm,
left=2mm,
right=2mm,
top=1.5mm,
bottom=1.5mm,
before skip=1mm,
after skip=2mm
]
\small\ttfamily\raggedright
\textbf{MATH and GSM8K:}\par
You are solving math problems. Separate each reasoning step with a blank line. Put the final answer in \textbackslash boxed\{\}.\par
\smallskip
\textbf{AMC and AIME:}\par
You are solving math competition problems. Separate each reasoning step with a blank line. Put the final answer in \textbackslash boxed\{\}.\par
\smallskip
\textbf{GPQA:}\par
You are solving a multiple-choice question. Separate each reasoning step with a blank line. Put the final answer letter in \textbackslash boxed\{\}.
\end{tcolorbox}

\begin{tcolorbox}[
enhanced,
breakable,
colback=gray!3,
colframe=SEUGreenDark,
colbacktitle=SEUGreenDark,
coltitle=white,
title={Prompt for LLM-guided local correction},
fonttitle=\bfseries,
boxrule=0.6pt,
arc=1mm,
width=\textwidth,
boxsep=1mm,
left=2mm,
right=2mm,
top=1.5mm,
bottom=1.5mm,
before skip=1mm,
after skip=2mm
]
\footnotesize\ttfamily\raggedright
You are a strict math competition teacher. Silently solve enough of the problem to judge the current step.\par
The candidate step may be mathematically wrong, incomplete, or going down an inefficient path.\par
Give concrete internal guidance for regenerating ONLY the current next step. Do not write the whole solution. If the current route is already valid, preserve it and make only the minimal correction needed.\par
Do not give vague advice such as ``consider using inequalities'' without naming the exact inequality, formula, or action.\par
If the candidate dropped a denominator, changed an equation incorrectly, made an unjustified assumption, or produced a premature final answer, explicitly correct that.\par
If this should not be the final step yet, say what exact operation should be done next instead of giving a final answer. Avoid changing to a different global method unless the existing route is truly invalid.\par
\smallskip
Original problem:\par
\{question\}\par
\smallskip
Accepted visible solution before the current step:\par
\{accepted\_text or [none]\}\par
\smallskip
Current no-thinking candidate:\par
\{current\_candidate\}\par
\smallskip
Earlier interventions for this same step:\par
\{history or [none]\}\par
\smallskip
Return only this format:\par
DIAGNOSIS: one sentence naming the exact issue or saying the step is acceptable.\par
NEXT\_STEP\_GUIDANCE: the concrete formula/calculation/reasoning move the small model should write for this current step.\par
MUST\_AVOID: one short warning about the main trap.
\end{tcolorbox}

\begin{tcolorbox}[
enhanced,
breakable,
colback=gray!3,
colframe=SEUGreenDark,
colbacktitle=SEUGreenDark,
coltitle=white,
title={Private correction note for guidance-conditioned regeneration},
fonttitle=\bfseries,
boxrule=0.6pt,
arc=1mm,
width=\textwidth,
boxsep=1mm,
left=2mm,
right=2mm,
top=1.5mm,
bottom=1.5mm,
before skip=1mm,
after skip=2mm
]
\small\ttfamily\raggedright
Private correction note:\par
\{guidance\}\par
\smallskip
Continue with only the next visible math step. Do not mention this note:
\end{tcolorbox}

\newpage

\section{Algorithm}
\label{Algorithm}
\vspace{-0.2cm}

\newcounter{rlcalgorithm} \refstepcounter{rlcalgorithm}
\label{alg:rlc}

\begin{tcolorbox}[
enhanced,
breakable,
colback=white,
colframe=SEUGreenDark,
colbacktitle=SEUGreenDark,
coltitle=white,
title={Algorithm \therlcalgorithm: Reusable Latent Correction (RLC)},
fonttitle=\bfseries,
boxrule=0.7pt,
arc=1mm,
width=\textwidth,
boxsep=1mm,
left=2mm,
right=2mm,
top=1.5mm,
bottom=1.5mm,
before skip=1mm,
after skip=2mm
]
\small
\noindent\textbf{Input:} offline corpus $\mathcal{D}_{\mathrm{off}}$, test problem $x$, frozen local model $f$, teacher LLM $g$, uncertainty threshold $\tau_u$, Router $R_\phi$, retrieval threshold $\tau_{\mathrm{sim}}$, top-$K$, and intervention weights $(\alpha_p,\alpha_r)$.\par
\noindent\textbf{Output:} answer $\hat{y}$ and an offline Experience Bank $\mathcal{B}$.\par
\smallskip

\noindent\textbf{Offline: Experience Bank construction}

\begin{enumerate}[leftmargin=2.5em,itemsep=0.2em,topsep=0.2em,parsep=0pt]
\item Initialize $\mathcal{B}\leftarrow\varnothing$.

\item \textbf{For each} $(x,y)\in\mathcal{D}_{\mathrm{off}}$, generate a stepwise Native trajectory with $f$ and maintain an accepted prefix.

\item At each step $t$, compute uncertainty $u_t$. If $u_t\leq\tau_u$, accept the Native step and continue.

\item Otherwise, call $g$ with the problem, accepted prefix, and candidate step to obtain localized guidance; privately condition $f$ on this guidance and regenerate only step $t$.

\item Extract key $k_t$ from the pre-error reasoning state. Construct $v_t^{\mathrm{proc}}$ from the unguided/guided prefix-state difference and $v_t^{\mathrm{result}}$ from the original/regenerated step-state difference. Add $(k_t,v_t^{\mathrm{proc}},v_t^{\mathrm{result}})$ to a pending set.

\item After the trajectory ends, commit its pending entries to $\mathcal{B}$ only if the final answer matches $y$.
\end{enumerate}

\smallskip
\noindent\textbf{Online: State-conditioned retrieval and intervention}

\begin{enumerate}[leftmargin=2.5em,itemsep=0.2em,topsep=0.2em,parsep=0pt]
\item Evaluate the problem-level route $r=\mathbb{I}\!\left[R_\phi(x)\geq 0.5\right]$. If $r=0$, return the Native answer generated by $f$.

\item If $r=1$, generate with $f$ one visible reasoning step at a time. Before generating each candidate step, form query state $q_t$ from the current reasoning prefix.

\item Compute the highest similarity $\max_i\rho_{t,i}$ between $q_t$ and the Bank keys. If $\max_i\rho_{t,i}<\tau_{\mathrm{sim}}$, accept the unmodified step and continue. Otherwise, retrieve the top-$K$ most similar Bank entries.

\item Aggregate the retrieved values using normalized similarity weights to obtain $(\bar{v}_t^{\mathrm{proc}},\bar{v}_t^{\mathrm{result}})$.

\item Add $\alpha_p\bar{v}_t^{\mathrm{proc}}$ to the final prefix-token state during prefill, and add $\alpha_r\bar{v}_t^{\mathrm{result}}$ to each newly decoded token state while regenerating the current step.

\item Append the corrected step to the visible reasoning trajectory and repeat until a final answer is produced; return $\hat{y}$.
\end{enumerate}
\end{tcolorbox}

\end{document}

%% file: math_commands.tex
\usepackage{amsmath,amsfonts,bm}

\def\eqref#1{equation~\ref{#1}}

\def\1{\bm{1}}

\DeclareMathAlphabet{\mathsfit}{\encodingdefault}{\sfdefault}{m}{sl}
\SetMathAlphabet{\mathsfit}{bold}{\encodingdefault}{\sfdefault}{bx}{n}

